\documentclass[10pt,twocolumn,letterpaper]{article}

\usepackage{cvpr}
\usepackage{times}
\usepackage{epsfig}
\usepackage{graphicx}
\usepackage{amsmath}
\usepackage{amssymb}

\usepackage{tabulary,multirow,xspace,xcolor} 
\usepackage{amsthm,fixmath,mathtools,nicefrac,mmstyle}

\usepackage[pagebackref=true,breaklinks=true,letterpaper=true,colorlinks,bookmarks=false]{hyperref}

\usepackage{bbm}
\usepackage{algorithm}
\usepackage{algpseudocode}
\usepackage{subfigure}
\usepackage{multicol}
\usepackage{enumitem}

\newenvironment{packed_enumerate}{
\vspace{-3pt}\begin{enumerate}
  \setlength{\itemsep}{0pt}
  \setlength{\parskip}{0pt}
  \setlength{\parsep}{0pt}
}{\end{enumerate}}

\cvprfinalcopy 

\def\cvprPaperID{1755} 

\ifcvprfinal\fi
\begin{document}

\title{EcoNAS: Finding Proxies for Economical Neural Architecture Search}

\author{Dongzhan Zhou$^1$\thanks{Equal contribution.} \quad Xinchi Zhou$^{1*}$ \quad Wenwei Zhang$^2$ \quad Chen Change Loy$^2$\\ Shuai Yi$^3$ \quad Xuesen Zhang$^3$ \quad Wanli Ouyang$^1$\\
$^{1}$The University of Sydney \hspace{10pt} $^{2}$Nanyang Technological University \hspace{10pt} $^{3}$SenseTime Research\\
{\tt\small \{d.zhou, xinchi.zhou1, wanli.ouyang\}@sydney.edu.au\hspace{10pt} \{wenwei001, ccloy\}@ntu.edu.sg} \\
{\tt\small \{yishuai, zhangxuesen\}@sensetime.com}
}

\maketitle
\thispagestyle{empty}

\begin{abstract}
Neural Architecture Search (NAS) achieves significant progress in many computer vision tasks. 
While many methods have been proposed to improve the efficiency of NAS, the search progress is still laborious because training and evaluating plausible architectures over large search space is time-consuming.
Assessing network candidates under a proxy (i.e., computationally reduced setting) thus becomes inevitable. 
In this paper, we observe that most existing proxies exhibit different behaviors in maintaining the rank consistency among network candidates. In particular, some proxies can be more reliable -- the rank of candidates does not differ much comparing their reduced setting performance and final performance.
In this paper, we systematically investigate some widely adopted reduction factors and report our observations.
Inspired by these observations, we present a reliable proxy and further formulate a hierarchical proxy strategy.
The strategy spends more computations on candidate networks that are potentially more accurate, while discards unpromising ones in early stage with a fast proxy. 
This leads to an economical evolutionary-based NAS (EcoNAS), which achieves an impressive 400$\times$ search time reduction in comparison to the evolutionary-based state of the art~\cite{Amoeba} (8 vs. 3150 GPU days).
{Some new proxies led by our observations can also be applied to accelerate other NAS methods while still able to discover good candidate networks with performance matching those found by previous proxy strategies.}
\end{abstract}

\section{Introduction}
Neural Architecture Search (NAS) has received wide attention and achieved significant progress in many computer vision tasks,
such as image classification~\cite{guo2019single, Amoeba, SNAS, TNAS}, detection~\cite{Chen2019DetNASBS, nas_fpn, wang2019nasfcos}, and semantic segmentation~\cite{Liu_2019_CVPR}.
Although recent NAS methods~\cite{PNAS, DARTS, ENAS} improve the search efficiency from earlier works~\cite{NAS},  
the search progress is still time-consuming and requires vast computation overhead when searching in a large search space since all network candidates need to be trained and evaluated.

\begin{figure}[t]
  \begin{center}
   \includegraphics[width=1.05\linewidth]{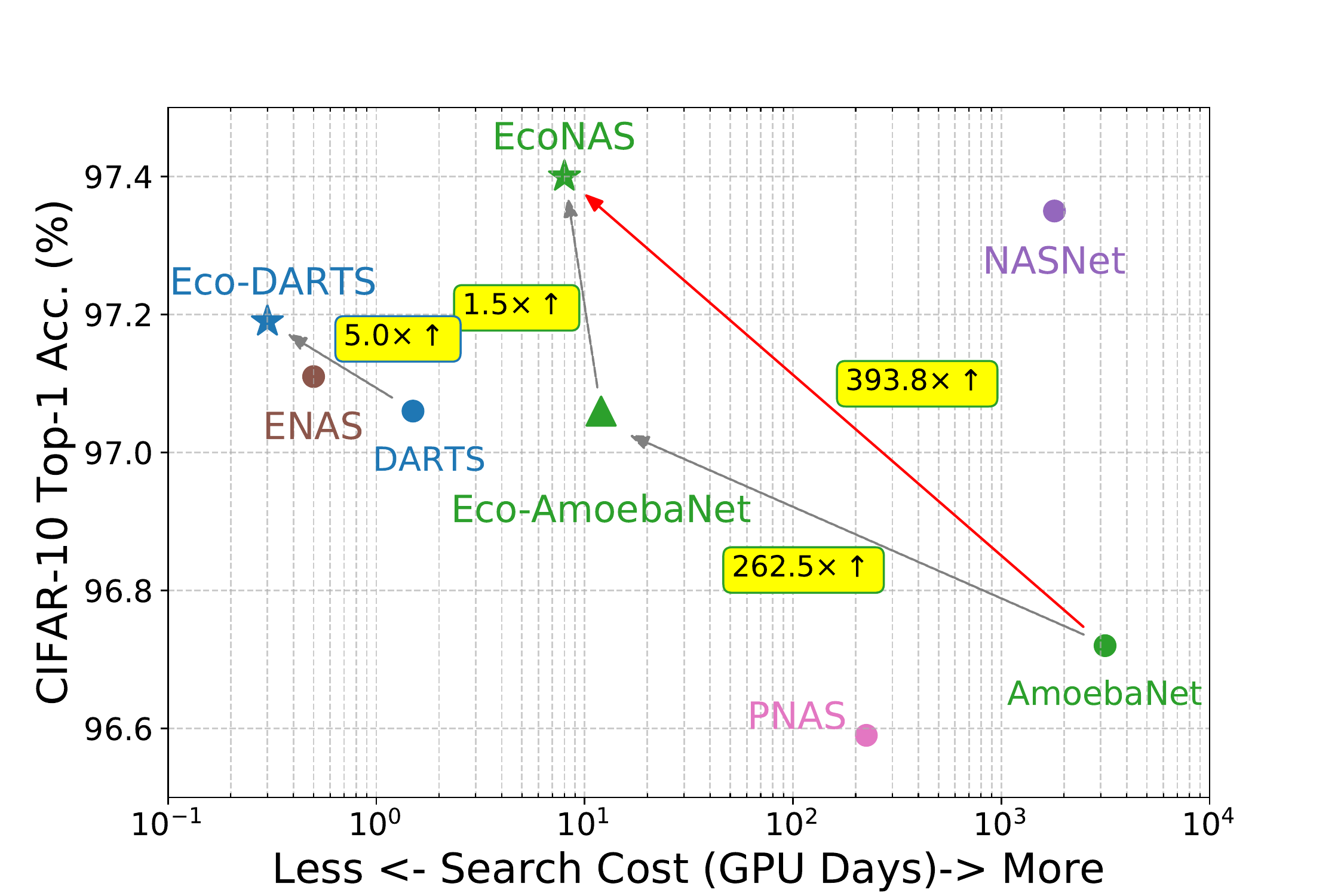}
  \end{center}
  \vspace{-9pt}
  \caption{
     The comparison of search cost and accuracy for different NAS methods on CIFAR-10~\cite{CIFAR10}.
     Simply replacing the original proxy with a more consistent and efficient one could reduce the search cost of DARTS~\cite{DARTS} and AmoebaNet~\cite{Amoeba}.
     The proposed EcoNAS uses an efficient proxy with a novel hierarchical proxy strategy, reducing around 400$\times$ search cost and achieving competitive performance comparing with AmoebaNet~\cite{Amoeba}.
  }
  \vspace{-12pt}
  \label{fig:brief}
\end{figure}

A widely adopted approach to alleviating this problem is by training and evaluating network candidates under proxies (\ie, computationally reduced settings~\cite{DARTS, ENAS, Amoeba, TNAS}). 
There are four common reduction factors, namely, the number of channels for CNNs ($c$), the resolution of input images ($r$), the number of training epochs ($e$), and the sample ratio ($s$) of the full training set.
These reduction factors either reduce the computation costs of networks or reduce the training iterations to save the search time.
While introducing proxies sounds appealing, a caveat is that different reduction factors exhibit different behaviors in keeping the rank consistency among network candidates. 
We observe that some architectures applied with certain reduction factors win in the reduced setting but perform worse in the original setup on CIFAR-10~\cite{CIFAR10}. 
Some other reduction factors perform more consistently. Those consistent and efficient reduced settings accelerate existing NAS methods and achieve competitive results as shown in Fig.~\ref{fig:brief}.

In this work, we investigate the behaviors of reduction factors $[c, r, s, e]$. Two main observations were obtained:
(1) with the same iteration numbers, using more training samples with fewer training epochs could be more effective than using more training epochs and fewer training samples in reducing the rank inconsistency; 
(2) reducing the resolution of input images is sometimes feasible while reducing the channels of networks is more reliable than reducing the resolution.
The aforementioned observations motivate us to design \textit{reliable proxies} that reduce channels and input resolutions while using all training samples. These new proxies apply well to many NAS methods, including evolutionary-based and gradient-based NAS methods~\cite{ProxylessNAS, DARTS, Amoeba}.
In particular, we observe consistent search time reduction when these new proxies are applied, while the discovered networks are similarly competitive to those found by the original NAS methods.

We further formulate a novel \textit{hierarchical proxy strategy} for evolutionary-based NAS.
The goal of the hierarchical proxy strategy is to discard less promising candidates earlier with a faster proxy and evaluate more promising candidates with a more expensive proxy.
The strategy is both effective and efficient: (1) this design saves substantial computation overhead by saving evaluation on less promising networks, and (2) assigning more resources to more promising networks help us to find good architectures more precisely.
Thanks to the hierarchical proxy strategy, the proposed \textit{economical evolutionary-based NAS (EcoNAS)} enjoys nearly 400$\times$ reduced search time (8 \vs 3150 GPU days) in comparison to the evolutionary-based state of the art~\cite{Amoeba} while maintaining comparable performance.

To summarize, our main contributions are as follows:
\begin{packed_enumerate}
\item We conduct extensive experiments to study commonly applied reduction factors. While inconsistency of reduction factors is well-known in the community, our work presents the first attempt to analyze the behaviour systematically. 
{\item Observations from the experiments lead to some fast and reliable proxies that are applicable to a wide range of NAS methods, including evolutionary-based NAS~\cite{Amoeba} and gradient-based NAS methods~\cite{ProxylessNAS, DARTS}. }
\item We present a hierarchical proxy strategy that leads to EcoNAS, reducing the search time requirement from thousands of GPU days to fewer than 10 GPU days. 
\end{packed_enumerate}
\section{Related Work}
\noindent\textbf{Neural Architecture Search.}
NAS aims at searching for good neural networks automatically in an elaborately designed search space.
One common approach of NAS is to train and evaluate each network candidates on a proxy, \ie, computation reduced setting, 
and to search architectures using either evolutionary algorithms (EA)~\cite{Baker2017AcceleratingNA, hier_evo, Amoeba, Real2017LargeScaleEO}, reinforcement learning (RL)~\cite{PNAS, ENAS, BlockQNN, NAS, TNAS}, or gradient-based methods~\cite{DARTS, SNAS}.
One-Shot methods~\cite{Bender2018UnderstandingAS, Brock2017SMASHOM, guo2019single} usually train a supernet covering the search space once
and then apply search algorithms to search the best path in this supernet as the searched architecture.

After the seminal work by Zoph and Le~\cite{NAS} that requires up to hundreds of GPU days to find a good architecture, many NAS methods~\cite{PNAS, DARTS, ENAS} try to reduce the search cost through different approaches.
The community first turns to searching for primary building cells rather than the entire network~\cite{BlockQNN, TNAS}.
Based on the cell search, some approaches try to use performance prediction based on learning curves~\cite{Baker2017AcceleratingNA} or surrogate models~\cite{PNAS} to expedite the evaluation process.
Parameter sharing between child models is also common for acceleration,\eg, in DARTS~\cite{DARTS}, ENAS~\cite{ENAS}, and One-Shot methods~\cite{Bender2018UnderstandingAS, Brock2017SMASHOM, guo2019single}.
Our EcoNAS is an evolutionary-based NAS method~\cite{Baker2017AcceleratingNA, hier_evo, Amoeba, Real2017LargeScaleEO}. 
Different from previous works that fix the proxies when searching for network architectures, we design a fast and consistent proxy to reduce the search time, and apply \emph{hierarchical proxy strategy} to improve the search efficiency.
It is noteworthy that another trend of NAS is to search for efficient architectures~\cite{ProxylessNAS, mobilenet_v3, tan2018mnasnet, FBNet} by adding constraints during search process such as latency~\cite{mobilenet_v3, tan2018mnasnet, FBNet} on specific platforms.
This differs to our focus on reducing the search cost by more careful and systematic proxy design.

\vspace{6pt}\noindent\textbf{Accelerating neural network training.}
Parallel computing~\cite{15_mins, large_bs_training, mixup} could significantly accelerate the training process of deep neural networks. 
Although these methods save the training time to minutes, the reduction in training time comes with the cost of thousands of GPUs, 
and their computation overhead remains large. 
Many studies~\cite{peephole, CNN_adv, All_Images} evaluate the performance of different networks at reduced settings, \eg, using smaller images in Tiny ImageNet Challenge\footnote[1]{\url{https://tiny-imagenet.herokuapp.com}}, 
using the encodings of network layers~\cite{peephole}, reducing the number of training iterations \cite{CNN_adv}, and reducing samples~\cite{All_Images}.
These studies assume that a specific reduced setting is sufficiently consistent thus do not evaluate the influence of different reduced settings. In contrast, we extensively evaluate influences from different reduced settings. 

\section{Exploration Study}\label{sec:exploration}

In this study, we investigate the behaviors of different reduced settings. Previous works have not studied such behaviours comprehensively and systematically.
%
%
To facilitate our study, we construct a model zoo containing 50 networks for comparing the rank consistency before and after applying reduced settings.
Each network in the model zoo is a stack of cells with alternating normal cells and reduction cells~\cite{PNAS, ENAS, TNAS}, generated by random sampling. 
The cell can be regarded as a directed acyclic graph consisting of an ordered sequence of nodes. 
Each node is a feature map in convolution networks and each directed edge is associated with an operation to be searched or randomly sampled.
The entire network structure and one example of the cell are shown in Fig.~\ref{fig:arch}(a) and Fig.~\ref{fig:arch}(c), respectively. 
The details are provided in the Appendix.

\begin{figure}
    \centering
    \includegraphics[width=0.48\textwidth]{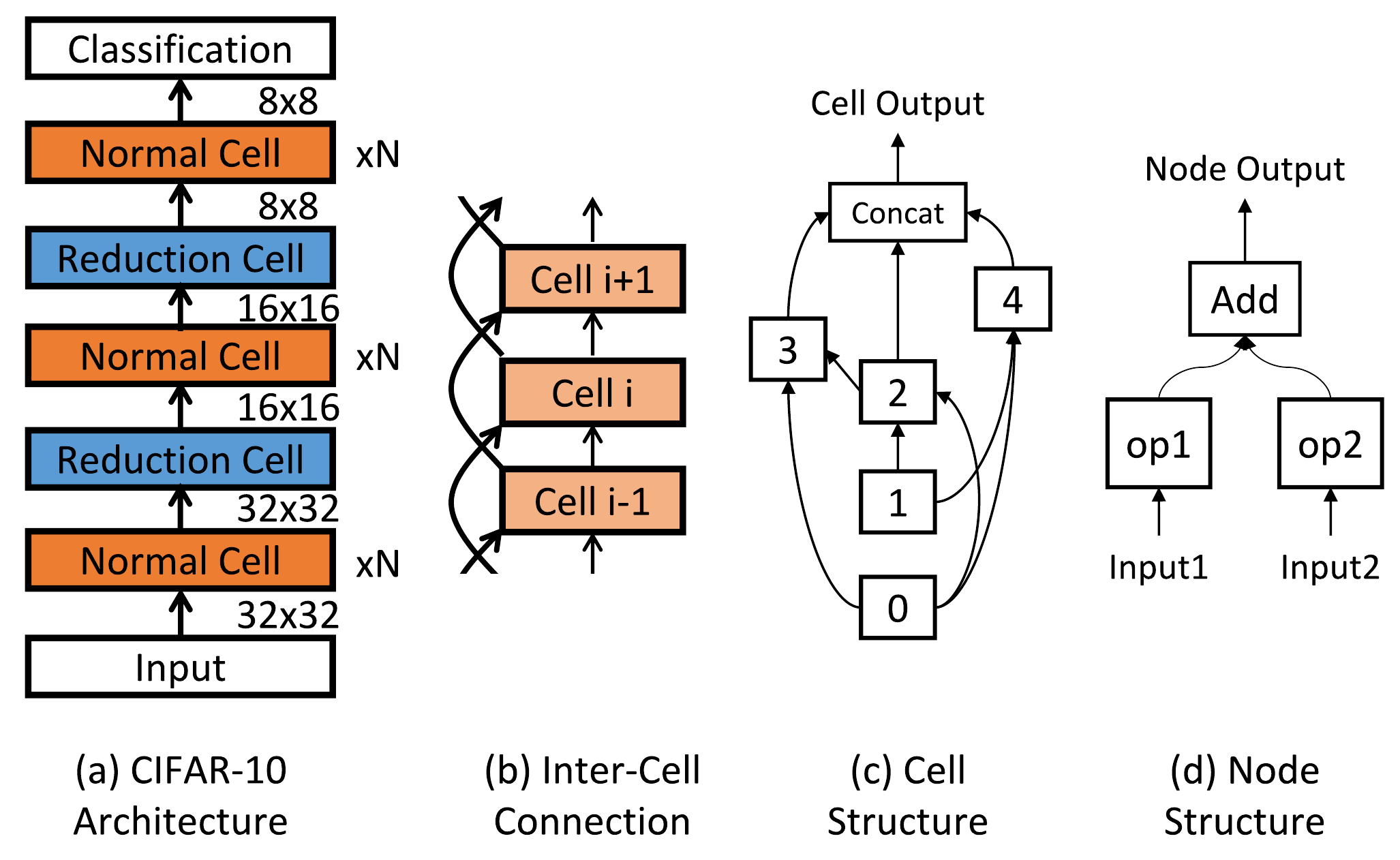}
    \vspace{-12pt}
    \caption{Network structure overview. (a) The cell stacking method for the whole network ($N=6$). (b) The connection between cells. (c) The general cell structure, where nodes 0, 1 are input nodes. (d) The construction of nodes 2, 3, 4 in the cell. }
    \label{fig:arch}
    \vspace{-9pt}
\end{figure}

\subsection{Reduction Factors}

In our experiments, a reduced setting corresponds to a combination of four factors: 
(1) number of channels for CNNs ($c$), (2) resolution of input images ($r$), 
(3) training epochs ($e$), and (4) sample ratio ($s$). 
%
The degree of reduction of $c$, $r$, and $s$ is represented by a subscript $0, 1, 2, 3, 4$. 
Specific values for the subscript of each reduction factor, for CIFAR-10 dataset~\cite{CIFAR10}, are shown in Table \ref{tab:red_cifar10}. 
The subscript for $e$ is an integer, \eg, $e_{30}$ indicates 30 training epochs. 
The value of $c$ refers to the number of channels in the initial convolutional cell. 
As for the sample ratio $s$, we randomly select a subset from the training data according to the value of $s$ and then train the network on this subset. 
%
For CIFAR-10, $(c_0, r_0, s_0, e_{600})$ corresponds to the original or conventional setting~\cite{DARTS}, which uses 36 initial channels and $32\times 32$ input image. All training samples are used for training with $600$ epochs. 
In comparison, $(c_a, r_b, s_c, e_{x})$ corresponds to a reduced setting requiring around $1/2^{a+b+c}$ FLOPs, \ie, $2^{a+b+c}$ speed-up, when compared with $(c_0, r_0, s_0, e_{x})$. 

\subsection{Evaluation Metric}
We use Spearman Coefficient, denoted as $\rho_{sp}$, as the metric to evaluate the reliability of reduced settings. %
The same metric is also used in \cite{PNAS} but it is employed for evaluating the performance of accuracy predictors. 
In our work, we use the metric to measure the dependence of ranks between the original setting and the reduced settings when the reduced settings are used for ranking the models in the model zoo. 
Assuming the model zoo has $K$ networks, the formulation of $\rho_{sp}$ is
\begin{equation}
\rho_{sp} = 1-\frac{6\sum_{i=1}^{K}d_i^2}{K(K^2-1)},
\label{eq_sp}
\end{equation}
where $d_i$ is the difference for network $i$ between its rank in the original setting and its rank in the reduced setting. 

\begin{figure}[t]
\subfigure[Small $\rho_{sp}$]{
\label{Fig2.sub.1}
\includegraphics[width=0.23\textwidth]{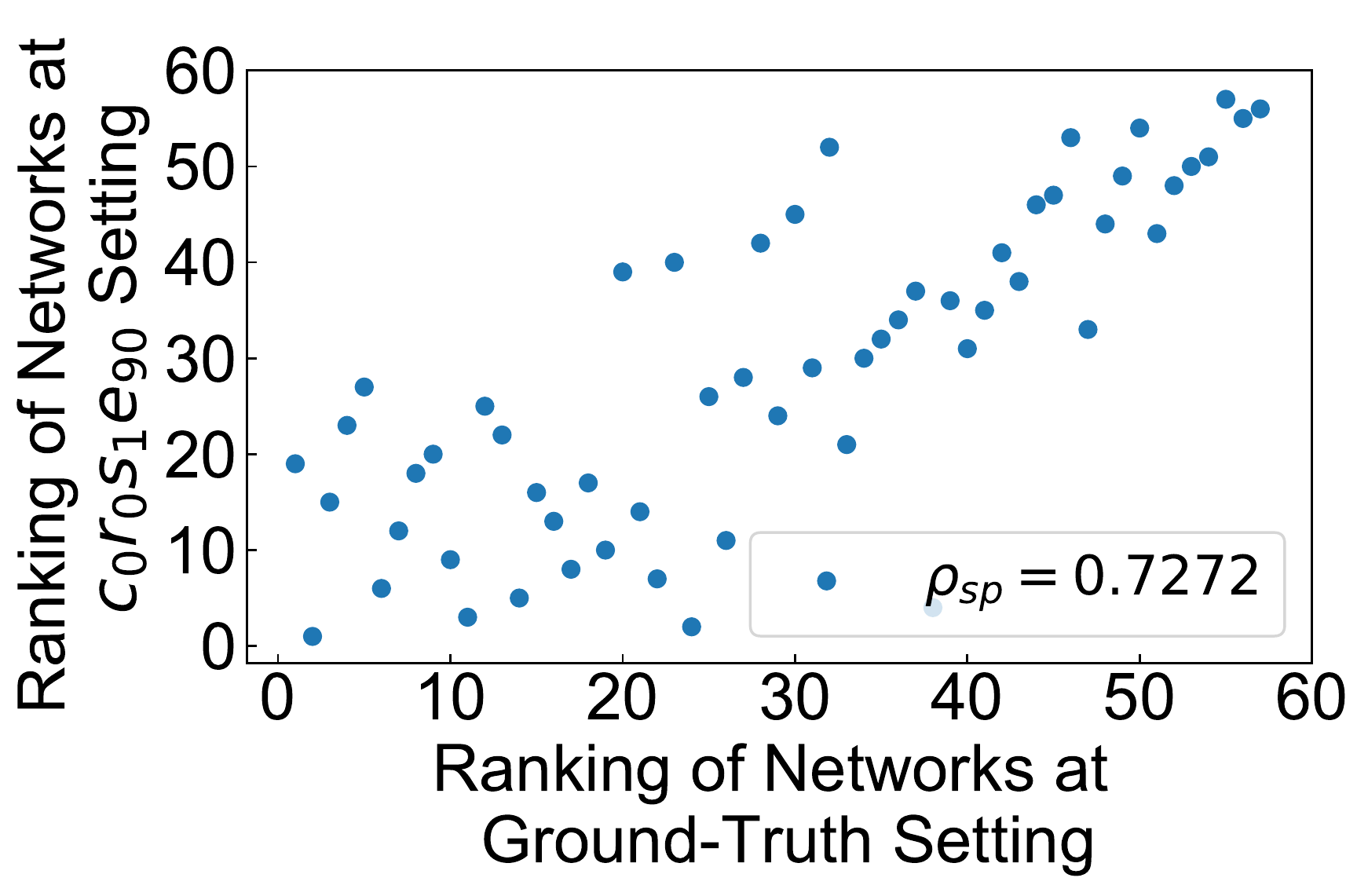}}
\subfigure[Large $\rho_{sp}$]{
\label{Fig2.sub.2}
\includegraphics[width=0.23\textwidth]{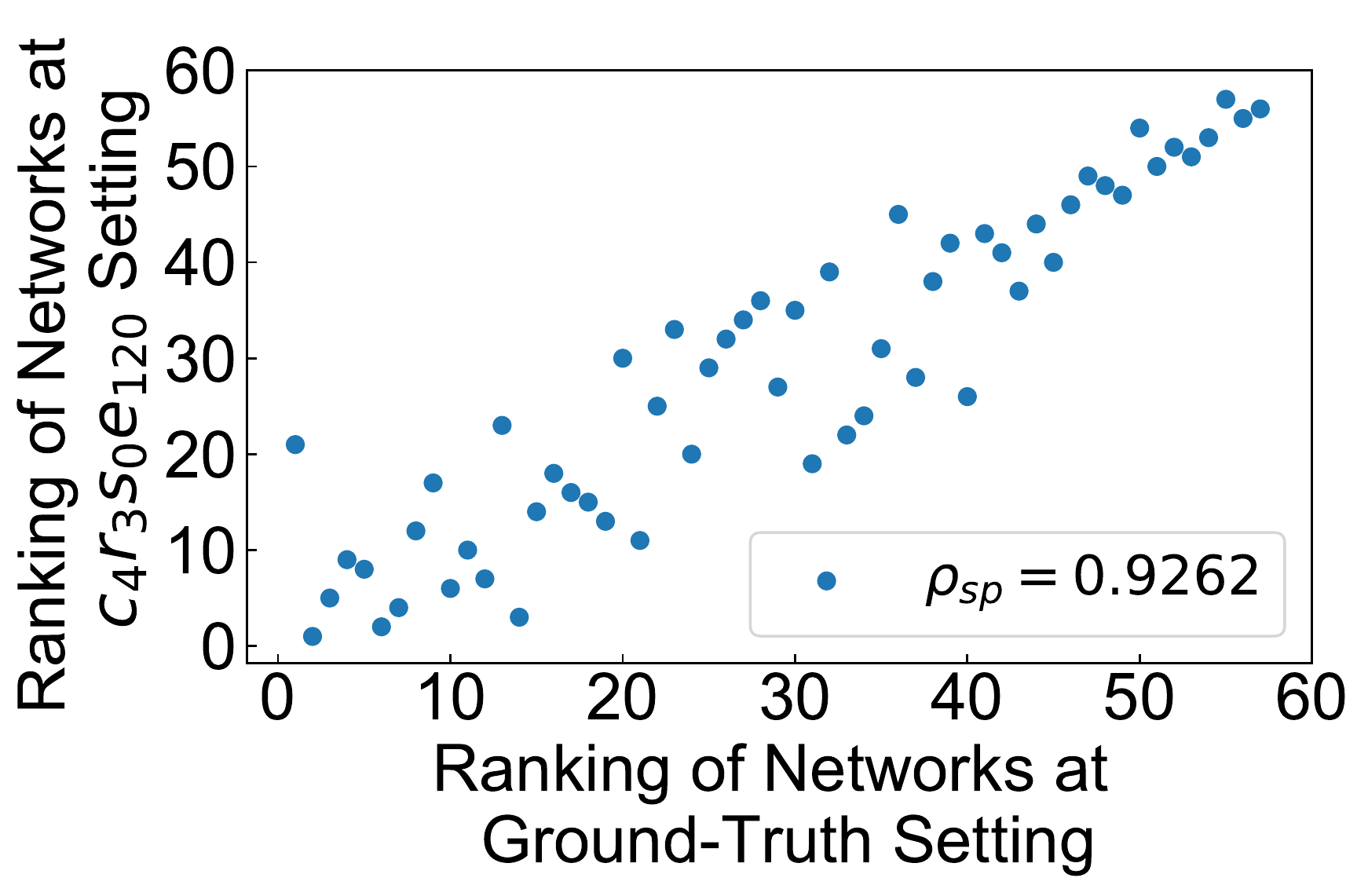}}
\vspace{-6pt}
\caption{
   Visualization of small $\rho_{sp}$ (a) and large $\rho_{sp}$ (b). 
   The $x$, $y$ coordinates of each point in the figure represent the ranks of a certain model in the original and reduced setting, respectively. 
   For instance, point (31,10) means that the rank of that model is 31 in the original setting but increases to 10 in the reduced setting.
}
\label{fig:sp_vis}
\vspace{-9pt}
\end{figure}

A higher Spearman Coefficient corresponds to a more reliable reduced setting. An architecture found in a reliable reduced setting is more likely to remain high rank in the original setting.
For example, Fig. \ref{fig:sp_vis} shows two different reduced settings, \ie, one with a small $\rho_{sp}$ (Fig.~\ref{Fig2.sub.1}) and another one with a large $\rho_{sp}$ (Fig.~\ref{Fig2.sub.2}) . 
When $\rho_{sp}$ is large, the dependence of ranks increases between the original setting and the reduced setting for the same model.
This indicates that the average change of ranks when switching from the reduced setting to the original one is small; thus, the reduced setting is more consistent and reliable.
When $\rho_{sp}$ is small, the rank obtained from the reduced setting is less reliable. 
The models with low ranks and high ranks at the original setting could rank highly and low at the reduced setting more frequently, respectively.
%

\begin{table}[t]
   \caption{
      Specific values of reduction factors for CIFAR-10~\cite{CIFAR10}. 
      For the values of $c_x$ and $r_x$, if $\frac{c_0(r_0)}{({1/\sqrt{2}})^x}$ ($x$ is the subscript) is an integer, the integer is taken directly ($c_2, r_2$). 
      Otherwise, the nearest number divisible by 4 is selected.
   }
   \vspace{-6pt}
   \small
   \begin{center}
   \begin{tabular}{|c|c|c|c|c|c|}
   \hline
   Reduction factor & 0 & 1 & 2 & 3 & 4 \\
   \hline
   $c$ & 36 & 24 & 18 & 12 & 8 \\
   $r$ & 32 & 24 & 16 & 12 & 8 \\
   $s$ & 1.0& 0.5&0.25&0.125& \\
   \hline\hline
   $e$ & 30 & 60 & 90 & 120& \\
   \hline
   \end{tabular}
   \end{center}
   \vspace{-18pt}
   \label{tab:red_cifar10}
\end{table}

\subsection{Experimental Observations}
\label{Sec:ExpCIFAR}
\begin{figure}[t]
\begin{center}
\includegraphics[width=0.3\textwidth]{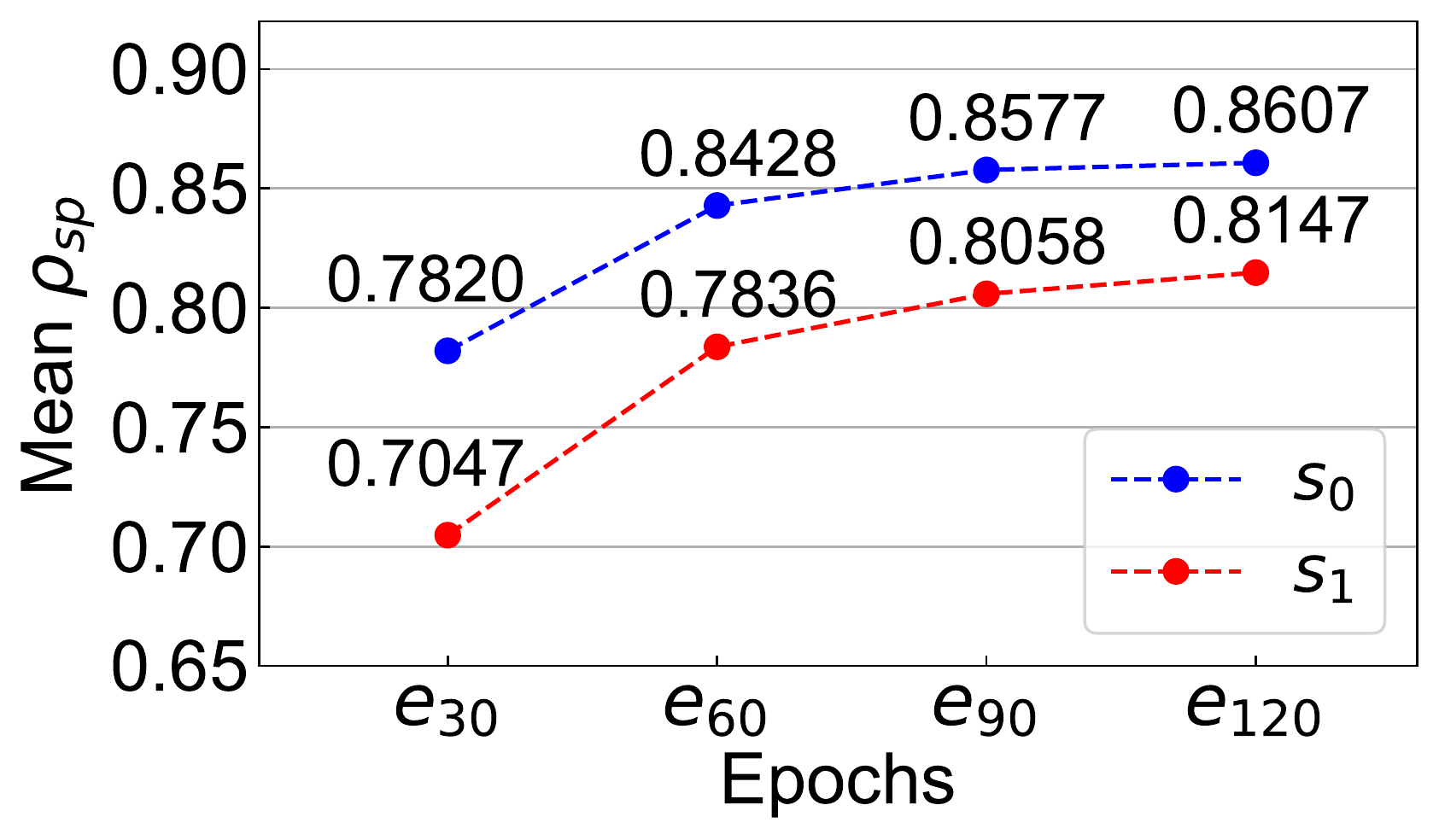}
\end{center}
\vspace{-12pt}
\caption{The influence of factors $s$ and $e$. The $Y$-axis is the average $\rho_{sp}$ for 25 different reduced settings that only differ in $c$ and $r$ but have the same $s$ and $e$.}
\label{fig:es}
\vspace{-12pt}
\end{figure}

We try different combinations of reduction factors and investigate the rank consistency between the original and reduced settings by Spearman Coefficient on the CIFAR-10 dataset~\cite{CIFAR10}.
The experimental setting is described in the Appendix.
Two useful observations are obtained as follows: 


\vspace{6pt}\noindent\textbf{1) With the same iteration numbers, using more training samples with fewer training epochs is more consistent than using more training epochs and fewer training samples.}
\label{se_section}
We first analyze the influence of sample ratio $s$ and training epoch $e$. 
There are 25 combinations of $c$ and $r$ for each combination of $s$ and $e$ as indicated in Table~\ref{tab:red_cifar10}.
As shown in Fig.~\ref{fig:es}, Spearman Coefficient $\rho_{sp}$ increases when the number of epochs and sample ratio increases. 
Therefore, the rank consistency improves with more training epochs and/or more training samples.  
The increase of $\rho_{sp}$ from 30 epochs to 60 epochs is the most obvious, after which the benefits brought about by more epochs become less apparent. 
When comparing the reduced setting pairs that have the same number of iterations such as $c_xr_ys_0e_z$ and $c_xr_ys_1e_{2z}$ where $x,y\in\{0,1,2,3,4\} ,z\in\{30,60\})$,
the results show that training with less epochs but using more samples in each epoch is a better choice than training with more epochs but fewer samples, especially when training for more iterations. 
%
Considering the trade-off between computation consumption and benefits, we find the combination $s_{0}e_{60}$ to be a more optimal setting than other combinations.

\begin{figure}[t]
   \begin{center}
      \includegraphics[width=0.65\linewidth]{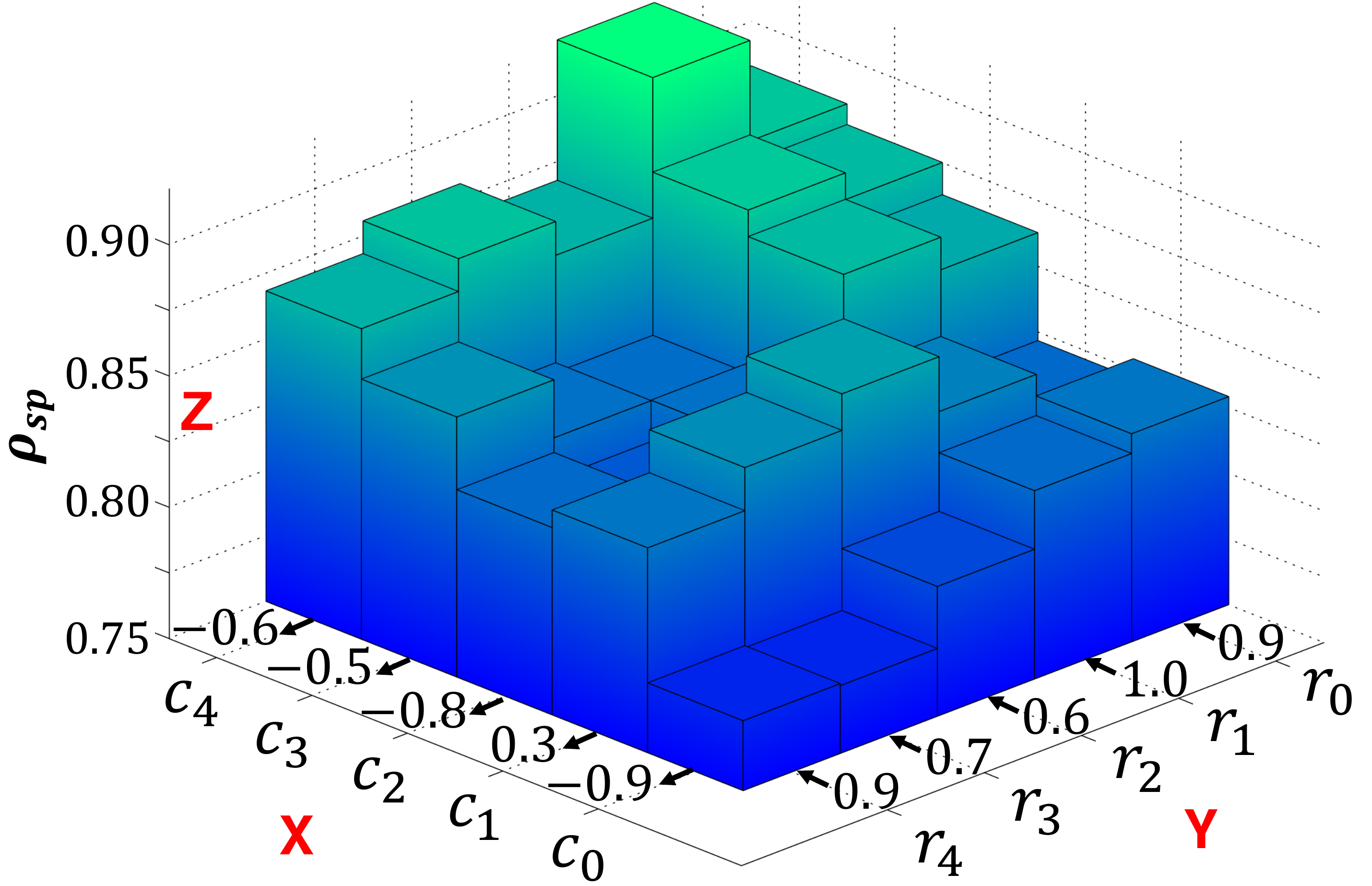}
   \end{center}
   \vspace{-6pt}
      \caption{3D-Bar chart for reduction factors $c$ and $r$. The $X$-$Y$ plane shows all the combinations of $c$ and $r$, and $Z$-axis represents the corresponding $\rho_{sp}$. We can directly compare $\rho_{sp}$ through the height of the bars. The values marked near the $X$ and $Y$ axes are $entropies$, which are used to evaluate monotonic increasing or decreasing trend for the $\rho_{sp}$ corresponding to the change of $r$ or $c$.}
   \vspace{-12pt}
   \label{fig:c_r}
\end{figure}

\begin{figure}[t]
   \subfigure[$c_xr_0s_0e_y$]{
   \label{Fig3.sub.1}
   \includegraphics[width=0.23\textwidth]{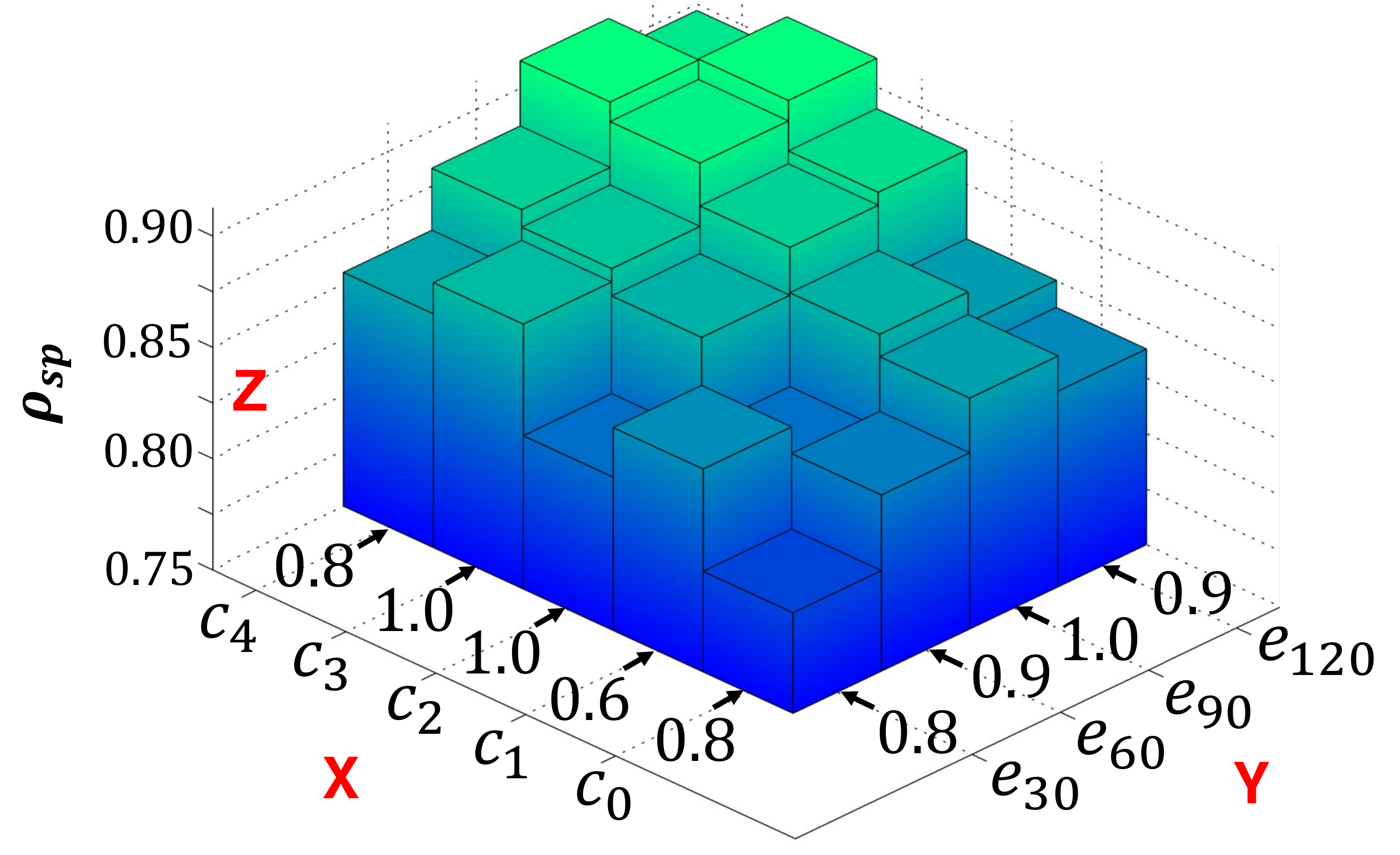}}
   \subfigure[$c_0r_xs_0e_y$]{
   \label{Fig3.sub.2}
   \includegraphics[width=0.23\textwidth]{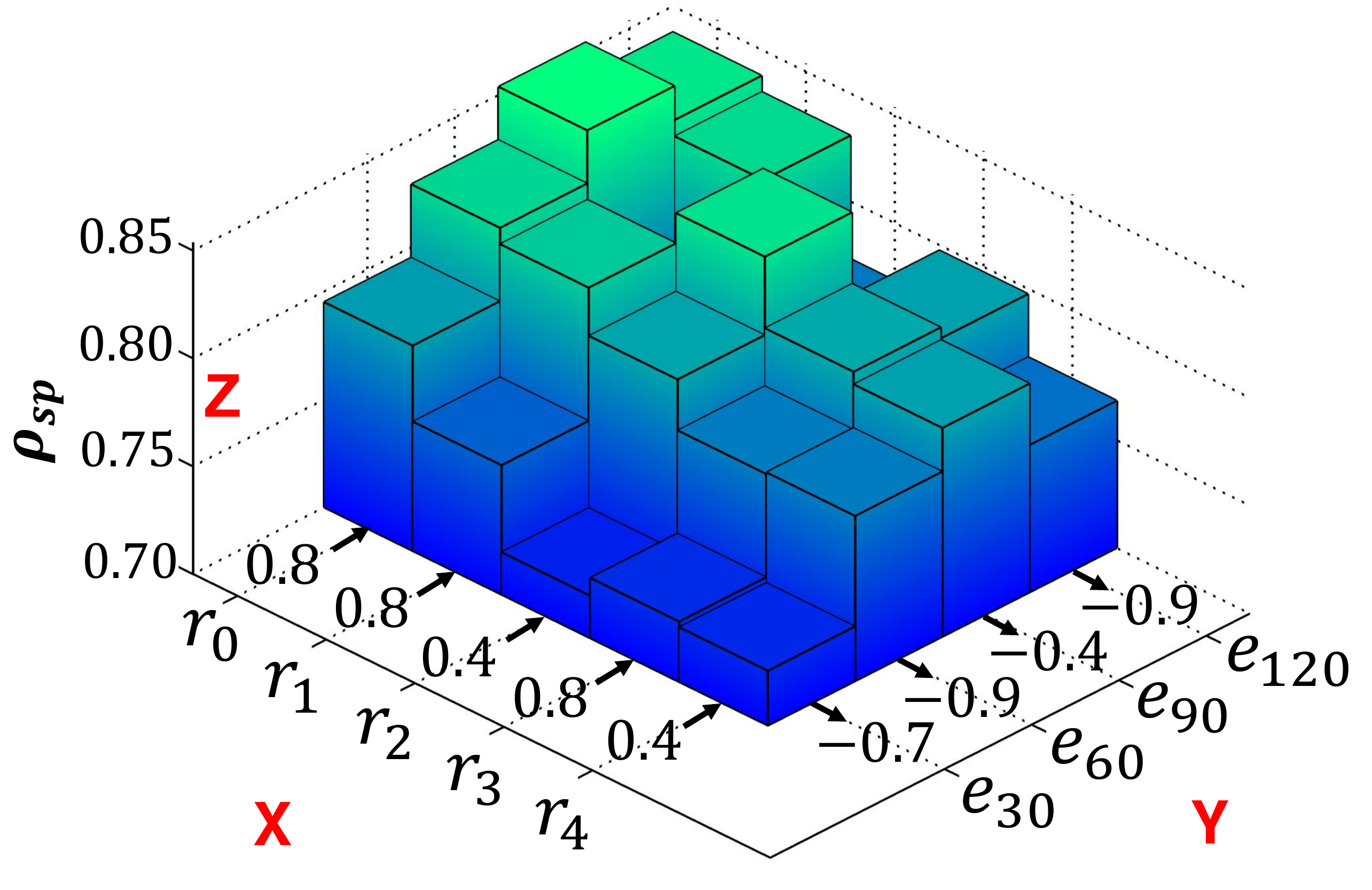}}
   \vspace{-6pt}
   \caption{3D-Bar chart of $c_xr_0s_0e_y$ (a) and $c_0r_xs_0e_y$(b), $x\in[0,1,2,3,4],y\in[30,60,90,120]$. We also show the values of $entropy$ of each dimension close to $X$ and $Y$ axes.}
   \label{fig:ce_re}
   \vspace{-9pt}
\end{figure}
   
 \begin{figure}[t]
    \begin{center}
    \includegraphics[width=0.8\linewidth]{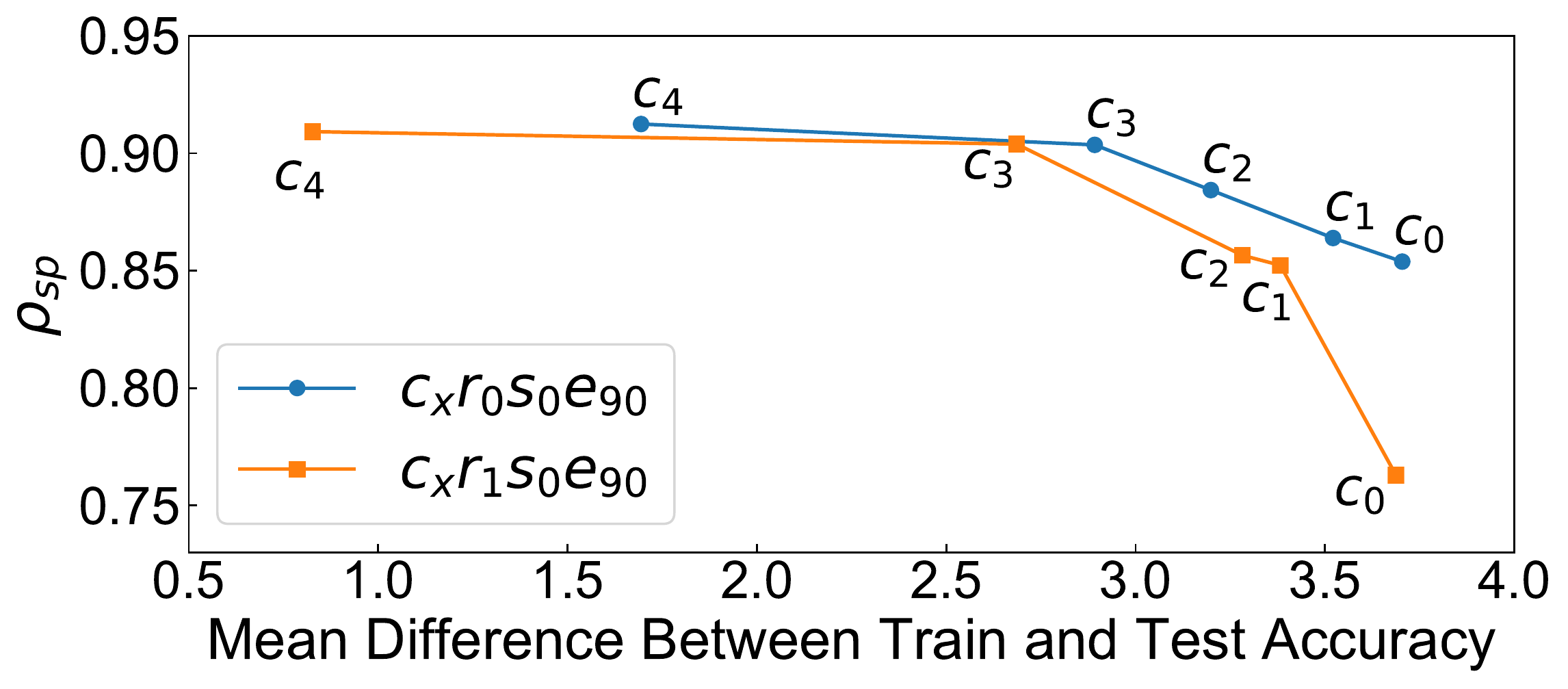}
    \end{center}
    \vspace{-12pt}
    \caption{The change of values $\rho_{sp}$ as the accuracy gap increases. 
    The $X$-axis is the mean difference of accuracy for networks in the model zoo between the train and test datasets at the end of training. 
    The difference evaluates the degree of overfitting under different reduced settings. 
    The $Y$-axis is the corresponding $\rho_{sp}$ for each reduced setting.
    }
    \vspace{-12pt}
    \label{fig:c_reason}
 \end{figure}

\vspace{6pt}\noindent\textbf{2) Reducing the resolution of input images is sometimes feasible while reducing the channels of networks is more reliable than reducing the resolution.}
\label{cr_section}
We further analyze the effect of reduction factors $c$ and $r$ with fixed $s_0e_{60}$ as discovered in our previous experiments. 
The 3D-Bar chart in Fig.~\ref{fig:c_r} shows the changes in $\rho_{sp}$ along the dimensions $c$ ($X$-axis) and $r$ ($Y$-axis). 
The $Z$-axis represents $\rho_{sp}$ for each $c_xr_y$ setting, illustrated by the height of the bars.  
We use a measurement called $entropy$ denoted by $\rho_e$ to indicate the monotonicity of $\rho_{sp}$ along a particular dimension of reduction factor. The details of this measurement are provided in the Appendix. 
The $entropy$ ranges from -1 to 1. The absolute value of $entropy$ close to 1 indicates that along this dimension, the objective set has an obvious monotonous increasing or decreasing trend, with only few small fluctuations. 
On the contrary, if the absolute value of $entropy$ is relatively small, \eg, less than 0.5, then the monotonic increasing or decreasing trend along this dimension is less apparent.


As shown by the values of $entropy$ close to the $Y$-axis in Fig.~\ref{fig:c_r}, the trend along the reduction factor $c$ is obvious. 
For most of the fixed $r$ settings, fewer channels lead to a better-behaved reduced setting, except for few points \eg, $c_2r_2$.
The variation of $entropy$ close to the $X$-axis indicates that the trends along $r$ for the fixed $c$ are not apparent. 
But for most of the cases, a smaller $r$ will lead to a smaller $\rho_{sp}$, which indicates that reducing the resolution is not reliable.

Figure~\ref{Fig3.sub.1} shows the effects of $c$ and $e$, where the other two factors are fixed to $r_0s_0$. 
The $entropies$ close to the $Y$-axis indicate that the decrease in $c$ will increase $\rho_{sp}$ for the same number of epochs, which is consistent with the results in Fig.~\ref{fig:c_r} for the same resolution. 
However, lower resolution leads to smaller $\rho_{sp}$, which is also consistent with the results in Fig.~\ref{fig:c_r}.
Both Fig.~\ref{Fig3.sub.1} and Fig.~\ref{Fig3.sub.2} show that more epochs improves $\rho_{sp}$, which also validates the results in Fig.~\ref{fig:es}. 

The results in Fig.~\ref{fig:c_r} and Fig.~\ref{Fig3.sub.1} indicate that decreasing the number of channels will lead to an increase of $\rho_{sp}$, which means that more reliable setting can be achieved by fewer channels with less computation and memory.
Fig.~\ref{fig:c_reason} also illustrates this phenomenon, where smaller accuracy difference appears with fewer channels between training and testing dataset.
%
This phenomenon may be caused by overfitting when the amount of parameters for the same architecture is reduced. 
We hypothesize that overfitting has an adverse effect on the rank consistency. 

\section{Economical Evolutionary-Based NAS}

Based on the investigations in the previous section, we propose \textbf{Economical evolutionary-based NAS (EcoNAS)}, an accelerated version of the evolutionary-based state of the art~\cite{Amoeba}. 
Existing evolutionary-based NAS methods~\cite{Baker2017AcceleratingNA, hier_evo, Amoeba, Real2017LargeScaleEO} suffer from large computation cost due to time wasted on  unpromising networks.
Taking heed of prior observations (Sec.~\ref{Sec:ExpCIFAR}), we designed a more efficient and consistent proxy (Sec.~\ref{sec:new_proxy}) that reduces the search cost by a large margin.
Then we propose a \emph{hierachical proxy strategy} that trains networks with different proxies based on their respective accuracy. 
This further improves the search efficiency by focusing more on good architectures and using the most accurate ones for evolution.
The search strategy and algorithm pipeline are described in Sec.~\ref{sec:algorithm}.

\subsection{Efficient Proxy}
\label{sec:new_proxy}
\begin{figure}[t]
	\begin{center}
	   \includegraphics[width=0.8\linewidth]{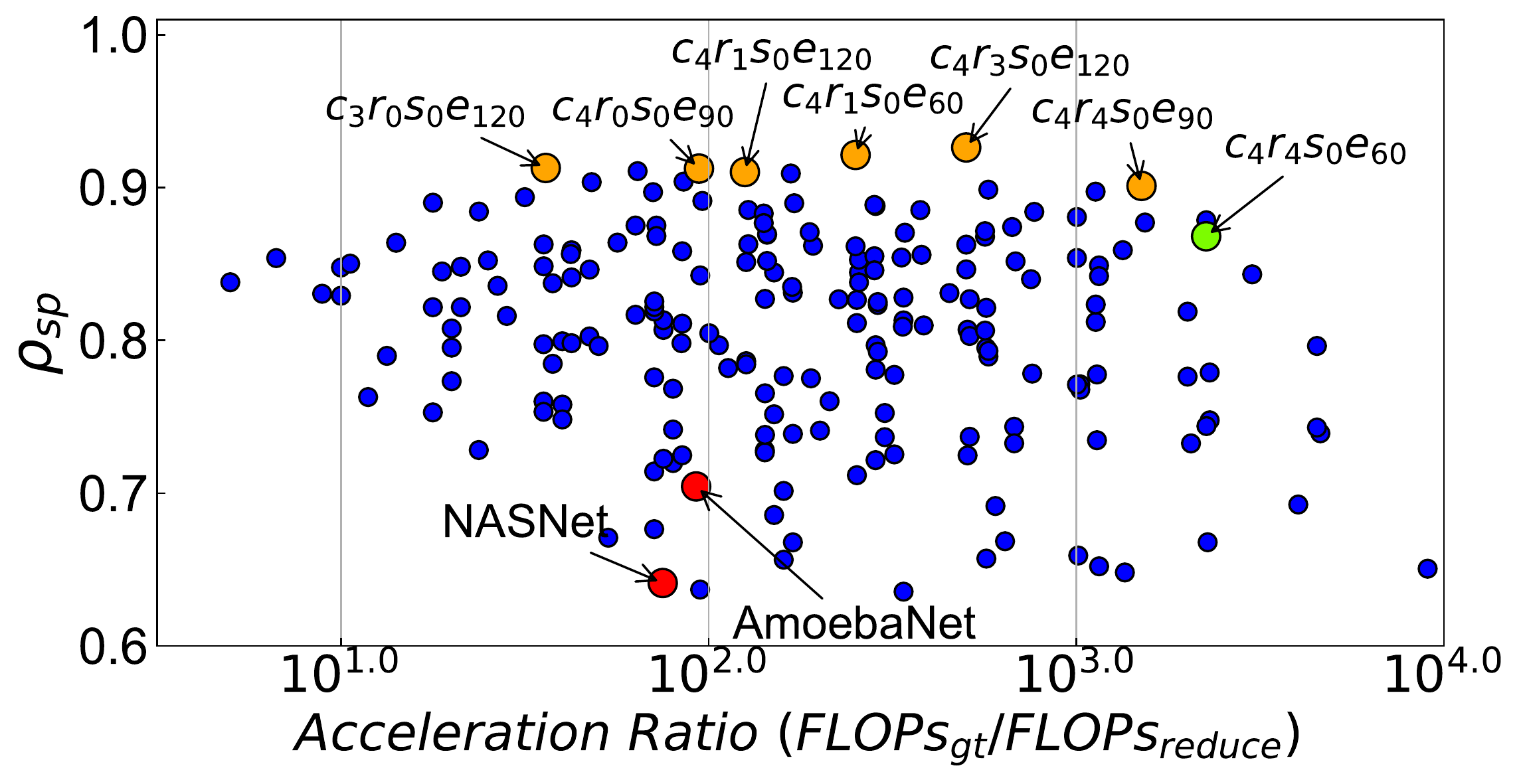}
	\end{center}
	\vspace{-12pt}
	   \caption{The rank consistency and acceleration ratios of different reduced settings on CIFAR-10~\cite{CIFAR10}. The $X$-axis represents the ratio of FLOPs at the original setting to that at the reduced setting. 
	   The $Y$-axis provides the corresponding $\rho_{sp}$ (Eq.~\eqref{eq_sp}). 
	   The reduced settings are getting more efficient along $X$-axis and getting more consistent in ranking different network candidates along $Y$-axis. 
	   The blue points show all the reduced settings, the orange points are good ones, and the green point is the recommended one.}
	\vspace{-9pt}
	\label{fig:recomm}
\end{figure}
To replace the original proxies in previous NAS methods~\cite{Amoeba, TNAS} with a more efficient one, we further analyze the acceleration ratio for FLOPs and $\rho_{sp}$ of the 200 reduced settings and provide empirical results on the good reduced settings. 
We divide the acceleration ratios into different groups and list the reduced settings that perform best in each group. 
We also compare two reduced settings in previous works~\cite{Amoeba, TNAS} where the stack number $N$ is set to 2. 
In NASNet~\cite{TNAS} the reduced setting is $c_0r_0s_0e_{20}$, and in AmoebaNet~\cite{Amoeba} the setting is $c_1r_0s_0e_{25}$. 
These two reduced settings exhibit less consistency as shown in Fig.~\ref{fig:recomm}.

Figure~\ref{fig:recomm} suggests that a proxy that substantially reduces the computation overhead does not necessarily have a poor rank consistency. 
There are many good reduced settings that possess large acceleration ratio and high consistency at the same time, just as the orange points in Fig.~\ref{fig:recomm}.
The orange points in Fig.~\ref{fig:recomm} show that many recommended reduced settings include $c_4$,  $e_{90}$, $e_{120}$. The observation verifies our previous conclusion about the benefits of more epochs and fewer channels. 
We adopt the reduced setting $c_4r_4s_0e_{60}$ in EcoNAS because this reduced setting exhibits relatively higher consistency and acceleration ratios as shown by the green point in Fig.~\ref{fig:recomm}. 
These reduced settings' exact values are shown in Table \ref{tab:red_cifar10}. 
Under this reduced setting, we enlarge the batch size from 96 to 384 to further compress the searching time.

\subsection{Hierarchical Proxy}
\label{sec:hierarchical_training}
%
Although we found an efficient and consistent proxy, 
we observe that training and evaluating each architecture with the same proxy still hurts the search efficiency because unpromising networks count a lot and waste most of the search time.
An intuitive approach to this problem is to reduce the training epochs further based on our newly designed consistent proxy.
This strategy could find and discard inaccurate architectures faster, however, the best model found from such a proxy with less training epochs might not be as good because the consistency will degrade given fewer training epochs.
Therefore, we propose \emph{hierarchical proxy strategy} as a trade-off to precisely rank good architectures and save time from assessing unpromising networks.

\vspace{6pt}\noindent\textbf{Population set.}
We divide the \emph{population} set into three subsets $P_{1}$, $P_{2}$, $P_{3}$, which contains networks with low, middle, and high accuracy and has small, middle, and high probability of being chosen for mutation, respectively.
For networks in the sets from $P_{1}$ to $P_{3}$, the networks are trained with faster but less consistent proxies (based on observations in Sec.~\ref{Sec:ExpCIFAR}).
We design these three proxies sharing the same $c$, $r$, $s$ but training networks with ${3E}$, ${2E}$, and ${E}$ epochs for $P_{3}$, $P_{2}$, and $P_{1}$ (\eg, $90$, $60$, and $30$ if $E=30$), respectively.
We apply this design because it exploits the weight sharing strategy and it is simple. For example, when several newly mutated networks remain after training $E$ epochs, we only need to train them with $E$ more epochs to rank them more precisely rather than to train with a new proxy from scratch.
More diverse proxies for the \emph{hierarchical proxy strategy} could also be tried and we leave it to future research.

During each evolution cycle, we use proxy with the lowest computation cost to quickly remove the less accurate networks from newly mutated networks.
Then the remaining networks, which are more accurate, will be evaluated with slower but more consistent proxy and assigned into the subsets $P_{2}$, $P_{3}$ hierarchically.
When choosing networks for mutation from the same subsets, the networks with a higher rank are more likely to be chosen to increase the probability of producing new accurate architectures.
Such a design saves the computation overhead by only training those networks with low accuracies using less epochs.
Furthermore, allocating resources to more promising architectures allows the good ones to be found with higher precsion according to the observation in Section~\ref{Sec:ExpCIFAR}.
Assigning larger probabilities to these networks for mutation also potentially helps to produce more promising architectures in fewer evolution cycles.

\vspace{6pt}\noindent\textbf{Algorithm pipeline.}
\label{sec:algorithm}
The pipeline of EcoNAS is shown in Algorithm \ref{Alg:Eco}.
\setlength{\textfloatsep}{6pt}
\begin{algorithm}[t]
	\small
	\begin{algorithmic}[1]
	\caption{EcoNAS Algorithm}
	\label{Alg:Eco}
	\State $P_{E} = \phi; P_{2E} = \phi; P_{3E} = \phi$ \\ \Comment{The population $P_k$ denotes networks trained for k epochs}
	\State $Train(model, a, b)$ \\ \Comment{The function that trains $model$, starting from epoch $a$, for totally $b-a$ epochs}
	\State $history=\phi$
	\While{$\rvert P_{E}\rvert< N_{init}$} \Comment{Initially, train $N_{init}$ models}
	\State{$model = $ RandomArchitecture$()$}
	\State{$model.accuracy = $ $Train(model,0,E)$} 
	\State{Add $model$ to $P_E$, $history$}
	\EndWhile
	\For{$cycle$ = $1$ to $C$ } \Comment{Evolve for $C$ cycles}
	
	\For {i = 1 to $N_0$}
	\State Randomly sample $model$ from $P_{E} ; P_{2E} ; P_{3E} $
	\State $child=$ RandomMutate$(model)$ 
	\State $child.accuracy=$ $Train(child,0,E)$ 
	\State Add $child$ to $P_E$, $history$
	\EndFor
	\For{model = top $1$ to $N_1$ models in $P_E$}
	\State $model.accuracy=$ $Train(model,E,2E)$ 
	\State Move $model$ from $P_E$ to $P_{2E}$
	\EndFor
	\For{model = top $1$ to $N_2$ models in $P_{2E}$}
	\State $model.accuracy=$ $Train(model,2E,3E)$ 
	\State Move $model$ from $P_{2E}$ to $P_{3E}$
	
	\EndFor
	\State Remove $dead$ from $P_{E} ; P_{2E} ; P_{3E} $
	\EndFor
	\end{algorithmic}
	\Return Several top models in $history$
\end{algorithm}
The initial architectures are randomly generated. All the searched architectures and their accuracies will be stored in the $history$ set. 
As explained before, in EcoNAS, there are three $population$ sets $P_{1}$, $P_{2}$, and $P_{3}$, also noted as $P_{E}$, $P_{2E}$, $P_{3E}$ because they store architectures that have been trained for $E$, $2E$, $3E$ epochs, respectively. 
For each evolution cycle, the following three steps are conducted: 

1. A batch of randomly sampled architectures from the population sets $P_E, P_{2E}$, and $P_{3E}$ will be mutated 
{(Algorithm~\ref{Alg:Eco}, line 12-17). Architectures with higher performance are more likely to be chosen. We follow the mutation rules in AmoebaNet~\cite{Amoeba} except that we remove the `identity' mutation, because in EcoNAS the amount of networks at each cycle is relatively fewer.} The mutated networks are trained from scratch for $E$ epochs and then added to set $P_{E}$. 

2. Choose the top architectures from set $P_E$ and $P_{2E}$ to $P_{2E}$ and $P_{3E}$, respectively (Algorithm~\ref{Alg:Eco}, line 18-25). These architectures are loaded from checkpoints and trained for $E$ more epochs, then those top architectures are more precisely ranked and moved to the corresponding subsets. 

3. Remove dead architectures from set $P_E$, $P_{2E}$ and $P_{3E}$ to force the algorithm to focus on newer architectures~\cite{Amoeba}.


\section{Experiments}

\noindent\textbf{Implementation Details}.
For EcoNAS, the number of nodes (Fig.~\ref{fig:arch}) within one architecture is four and we set the stack number $N=6$ to build the networks. 
We found in our experiments that $N=2$, a setting used in previous works~\cite{Amoeba, TNAS} is detrimental to the rank consistency (Fig.~\ref{fig:recomm}).
%
Our searching experiments are conducted on CIFAR-10~\cite{CIFAR10} and we split 5k images from the training set as validation set. 
The search space is similar to that in \cite{DARTS}.
Initially, $P_1$ has 50 randomly generated architectures, while $P_2$ and $P_3$ are empty. At each cycle, 16 architectures sampled from $P_k,k\in[E,2E,3E]$ will be mutated. We set the number of cycles $C$ to 100 and training epoch $E$ to 20.
All the experiments and search cost calculation are conducted using NVIDIA GTX 1080Ti GPUs.

\subsection{Overall Results on CIFAR-10 and ImageNet}\label{sec:cifar_imagenet}
The searching process yields several candidate convolutional cells and these cells are evaluated on two datasets, CIFAR-10~\cite{CIFAR10} and ImageNet~\cite{ILSVRC15} for image classification tasks. 
Unlike previous works~\cite{Amoeba, BlockQNN}, which select dozens or hundreds of cells to evaluate, we only pick up the top-5 cells from $history$ to evaluate the performance of our search algorithm.
Retraining top-5 cells saves the retraining overhead and is enough in our experiments; the reason is analyzed in Section~\ref{sec:analysis}.

\vspace{6pt}\noindent\textbf{Results on CIFAR-10.}
For the task of CIFAR-10, we set $N=6$ and the initial channel $c=36$. 
The networks are trained from scratch for 600 epochs with batch size 96. 
Other hyper-parameters follow previous works \cite{DARTS, SNAS}.
The experimental results on CIFAR-10 are shown in Table \ref{results:cifar10}. 
The best model found by our algorithm achieves a test error of 2.62\%, a rate that is on par with state-of-the-art evolution-based method and is much lower than most Reinforcement Learning (RL)-based methods.
Importantly, the model uses about $400\times$ less computation resources. 
Under the same magnitude of computational complexity, our result is superior to gradient-based methods~\cite{DARTS, SNAS} and weight sharing method~\cite{ENAS}.

\begin{table}
	\small\caption{CIFAR-10 test errors of EcoNAS and the state-of-the-art networks. The `Error' refers to top-1 error rate and `Cost' refers to the number of GPU days.}
	 \label{results:cifar10}
	 \vspace{-6pt}
 	\begin{center}
	\small
	\addtolength\tabcolsep{-0.3em}
 	\begin{tabular}{cccc}
		\hline
		Network & Error (\%) & Params. (M) & Cost \\
		\hline
		DenseNet-BC\cite{DenseNet} & 3.46 & 25.6 & - \\
		\hline
		NASNet-A \cite{TNAS} & 2.65 & 3.3 & 1800 \\
		Amoeba-A \cite{Amoeba} & 3.34$\pm$0.06 & 3.2 & 3150 \\
		Amoeba-B \cite{Amoeba} & 2.55$\pm$0.05 & 2.8 & 3150(TPU) \\
		Hierarchical Evo \cite{hier_evo} & 3.75$\pm$0.12 & 15.7 & 300 \\
		PNAS\cite{PNAS} & 3.41$\pm$0.09 & 3.2 &225 \\
		ENAS\cite{ENAS} & 2.89 & 4.6 & 0.5 \\
		\hline
		DARTS(1st order)\cite{DARTS} & 3.00$\pm$0.14 & 3.2 & 1.5 \\
		DARTS(2nd order)\cite{DARTS} & 2.76$\pm$0.09 & 3.3 & 4 \\
		SNAS\cite{SNAS} & 2.85$\pm$0.02 & 2.8 & 1.5 \\
		\hline
		EcoNAS & \textbf{2.62$\pm$0.02} & 2.9 & 8 \\
		\hline
	 \end{tabular}
	 \end{center}
	 \vspace{-12pt}
\end{table}

\vspace{6pt}\noindent\textbf{Results on ImageNet.}
To evaluate the transferability of the cells searched by EcoNAS, we transfer the architectures to ImageNet, where we only change the stack number $N$ to 14 and enlarge the initial channel $c$ to 48. 
As shown in Table~\ref{results:imnet}, the best architecture found by EcoNAS on CIFAR-10 generalizes well to ImageNet.
EcoNAS achieves top-1 error that outperforms the previous works that consume the same magnitude of GPUs~\cite{DARTS, SNAS}. 
EcoNAS also surpasses models with similar amount of parameters (\ie, having fewer than 6M parameters) found by reinforcement learning~\cite{TNAS} and evolution algorithms~\cite{Amoeba}, which require about 200$\times$ more GPU resources.

\begin{table}
	\small\caption{ImageNet Test errors of EcoNAS and the state-of-the-art networks. The `Error' refers to top-1 error rate and `Cost' refers to the number of  GPU days.}
	\label{results:imnet}
	\vspace{-6pt}
	\begin{center}
 	\begin{tabular}{cccc}
		\hline
		Architecture & Error (\%)& Params. (M) & Cost \\
		\hline
		NASNet-A\cite{TNAS} & 26.0 & 5.3 & 1800 \\
		NASNet-B\cite{TNAS} & 27.2 & 5.3 & 1800 \\
		NASNet-C\cite{TNAS} & 27.5 & 4.9 & 1800 \\
		AmoebaNet-A\cite{Amoeba} & 25.5 & 5.1 & 3150 \\
		AmoebaNet-B\cite{Amoeba} & 26.0 & 5.3 & 3150 \\
		AmoebaNet-C\cite{Amoeba} & 24.3 & 6.4 & 3150 \\
		DARTS\cite{DARTS} & 26.9 & 4.9 & 4 \\
		SNAS\cite{SNAS} & 27.3 & 4.3 & 1.5 \\
		\hline
		EcoNAS & \textbf{25.2} & 4.3 & 8 \\
		\hline
	 \end{tabular}
	 \end{center}
	 \vspace{-3pt}
\end{table}

\subsection{Ablation Study of EcoNAS}\label{sec:ablation}
\noindent\textbf{Reduced setting for NAS using evolutionary algorithms.}
We evaluate the reduced setting on the conventional evolutionary algorithms in Table~\ref{tab:econas_ablation}. AmoebaNet~\cite{Amoeba}, NASNet~\cite{TNAS}, and our EcoNAS use the same search space but they have different in search algorithm design.
By simply replacing the reduced setting of AmoebaNet~\cite{Amoeba} to more consistent ones, its computational costs are remarkably reduced and the accuracies of searched models also increase, with and without applying \emph{hierarchical proxy strategy}. 

\vspace{6pt}\noindent\textbf{Reduced setting for other NAS methods.}
We further evaluate the effectiveness of the newly discovered reduced settings on other NAS methods, \eg, gradient-based methods~\cite{ProxylessNAS, DARTS}. 
We report the top-1 error rates of DARTS~\cite{DARTS} on CIFAR-10~\cite{CIFAR10}
and ProxylessNAS~\cite{ProxylessNAS} on ImageNet-1k~\cite{ILSVRC15}, as shown in Table \ref{tab:reduced_setting}. 
{Directly reducing the channels and the input resolutions by half accelerates these methods and finds models that keep comparable accuracy in comparison to those searched by the original proxies.
This validates our observation reported in Sec.~\ref{Sec:ExpCIFAR} and the effectiveness of our newly designed reduced setting.}

\begin{table}[tbp]
	\small\caption{
		Ablation study of EcoNAS. Reduced setting $c_1r_0s_0e_{25}$ is used in AmoebaNet~\cite{Amoeba} and $c_0r_0s_0e_{20}$ is used in NASNet~\cite{TNAS}. We use $c_4r_4s_0e_{60}$ in EcoNAS.
		`H.P.' denotes whether our proposed \emph{hierarchical proxy strategy} is used. `Cost' denotes the number of GPU days used. `Sp' denotes the Spearman Coefficient. `Error' refers to top-1 error rate.}
	\label{tab:econas_ablation}
	\vspace{-6pt}
	\small
	\addtolength\tabcolsep{-0.4em}
	\begin{center}
    \begin{tabular}{c|c|c|c|c|c}
    \hline
    Reduced Setting & H.P.& Cost & Sp & Params. (M)& Error (\%) \\
	\hline
	AmoebaNet~\cite{Amoeba} & & 3150 & 0.70 & 3.2 & $3.34$ \\
	$c_4r_4s_0e_{35}$ (Ours) & & 12 & 0.74 & 3.2 &$2.94$ \\
	\hline
	NASNet~\cite{TNAS}& \checkmark & 21 & 0.65 & 2.9 & $3.20$ \\
   $c_3r_2s_1e_{60}$ & \checkmark & 12 & 0.79 & 2.6 & $2.85$ \\
    $c_4r_4s_0e_{60}$ (Ours) & \checkmark & 8 & 0.85 & 2.9 & $2.62$ \\
    \hline
	\end{tabular}
	\end{center}
	\vspace{-12pt}
\end{table}

\begin{table}[t]
	\small\caption{Results of applying new reduced settings to other NAS methods. 
	`Error' indicates top-1 error rate of DARTS on CIFAR-10~\cite{CIFAR10} and of ProxylessNAS on ImageNet-1k~\cite{ILSVRC15}.
	`Cost' indicates the number of GPU days used.}
	\label{tab:reduced_setting}
	\vspace{-6pt}
	\addtolength\tabcolsep{-0.5em}
	\small
    \begin{center}
    \begin{tabular}{c|c|c|c|c}
    \hline
       Method & Setup & Cost & Params. (M) & Error (\%) \\
	\hline
		\multirow{3}{*}{DARTS~\cite{DARTS}}
        &$c_2r_0s_0$(1st)\cite{DARTS} & 1.5 & 3.2 & $3.00$ \\ 
        & $c_2r_0s_0$(2nd)\cite{DARTS}& 4   & 3.3 & $2.76$ \\
		& $c_4r_2s_0$(Ours)          & 0.3 & 4.5 & $2.80$ \\
		\hline
		\multirow{3}{*}{ProxylessNAS~\cite{ProxylessNAS}}
        &$c_0r_0s_0$-S ~\cite{ProxylessNAS} & 8  & 4.1 & $25.4$ \\ 
        &$c_0r_0s_0$-L ~\cite{ProxylessNAS} & 8  & 6.9 & $23.3$ \\ 
        &$c_2r_2s_0$ (Ours)               & 4  & 5.3 & $23.2$ \\
    \hline
	\end{tabular}
	\end{center}
	\vspace{-3pt}
\end{table}

\vspace{6pt}\noindent\textbf{Hierarchical proxy strategy.}
We evaluate the \emph{hierarchical proxy strategy} on the conventional evolutionary method~\cite{Amoeba} in Table~\ref{tab:econas_ablation}. 
{The reduced settings $c_4r_4s_0e_{60}$ and $c_4r_4s_0e_{35}$ train the same number of models (1k) for the same number of epochs (35k) with or without the \emph{hierarchical proxy strategy}, respectively.
Both the search cost and error rate are reduced after applying the \emph{hierarchical proxy strategy}.
The results suggest the search efficiency of \emph{hierarchical proxy strategy} and its effectiveness in finding better networks using less search time.
}

\subsection{Analysis}\label{sec:analysis}
\begin{figure}[t]
	\begin{center}
	   \includegraphics[width=0.8\linewidth]{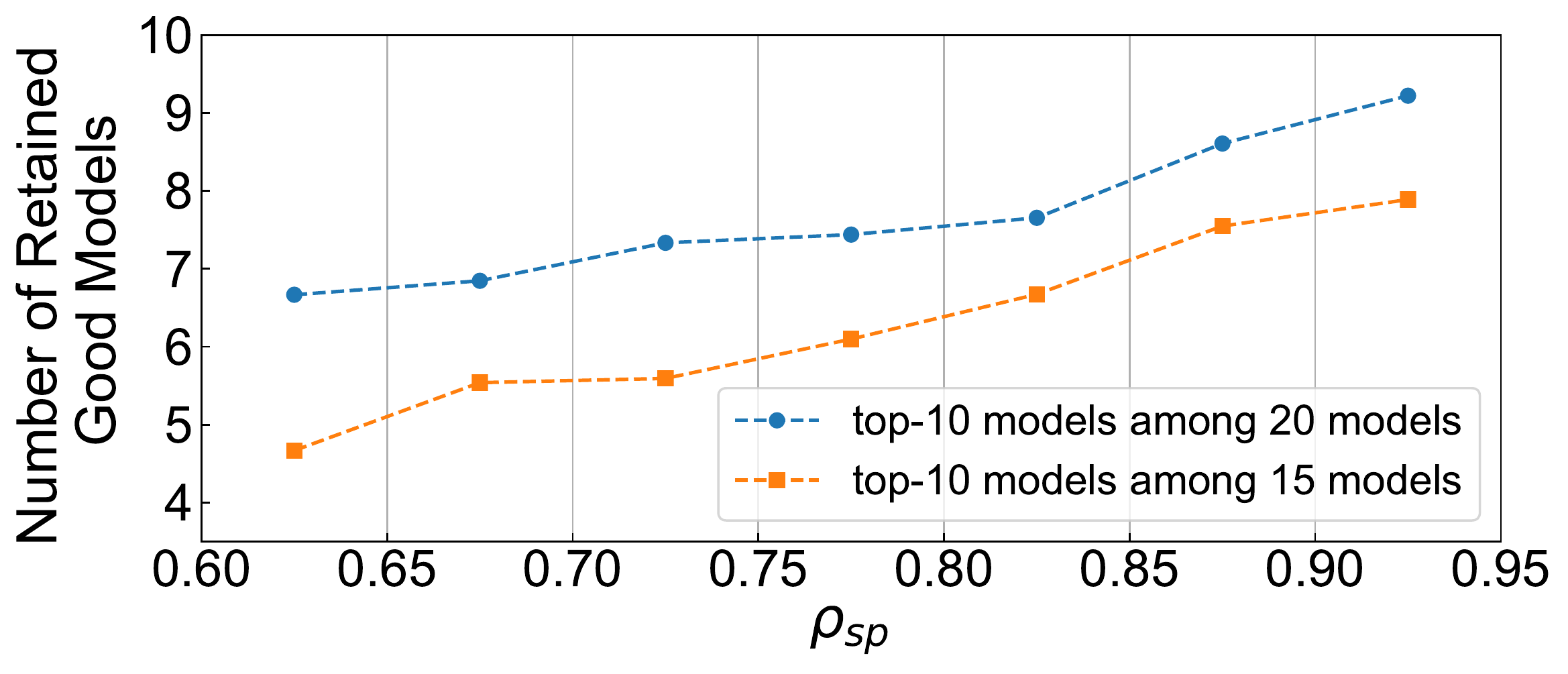}
	\end{center}
	\vspace{-9pt}
	   \small\caption{Ability to retain top models for reduced settings with different $\rho_{sp}$.}
	\vspace{6pt}
	\label{fig:retain}
\end{figure}

\noindent\textbf{Ability of reliable proxies to retain top models.}
We further evaluate the ability of the reduced settings in retaining good models. 
If a top-10 model at the original setting is a top-15 or top-20 model at the reduced setting, it will be considered as a `good model retained'.
We divide the reduced settings into different groups according to their corresponding $\rho_{sp}$ and calculate the average number of good models retained for reduced settings in the same group.
As $\rho_{sp}$ increases, the number of good models retained also increases (Fig.~\ref{fig:retain}), 
suggesting that reduced settings with better consistency can keep more good models in the top positions.

Figure~\ref{fig:retain} also indicates that \emph{reliable proxies can reduce not only the search cost but also the retraining overhead}.
Previous works need to select the best model after retraining hundreds of top models searched under proxies (\eg, 100~\cite{BlockQNN} for BlockQNN and 250 for NASNet~\cite{TNAS}),
AmoebaNet~\cite{Amoeba} also needs to retrain 20 models.
One of the reasons for this phenomenon is that previous works adopt less reliable proxies as shown in Fig.~\ref{fig:recomm}.
Since the ranks of networks under the proxies are less consistent with their actual ranks under the original setting, they need to select more networks for retraining to find the optimal one.
According to the conclusion obtained from Fig.~\ref{fig:retain}, a more consistent reduced setting retains more top models, 
thus allows the search algorithm to retrain fewer networks to obtain competitive accuracy.
With a more reliable proxy, EcoNAS only retrains top-5 networks and remarkably saves computation overhead for retraining models, which is usually overlooked in most NAS literature. 

\begin{figure}[t]
	\begin{center}
	  \includegraphics[width=0.7\linewidth]{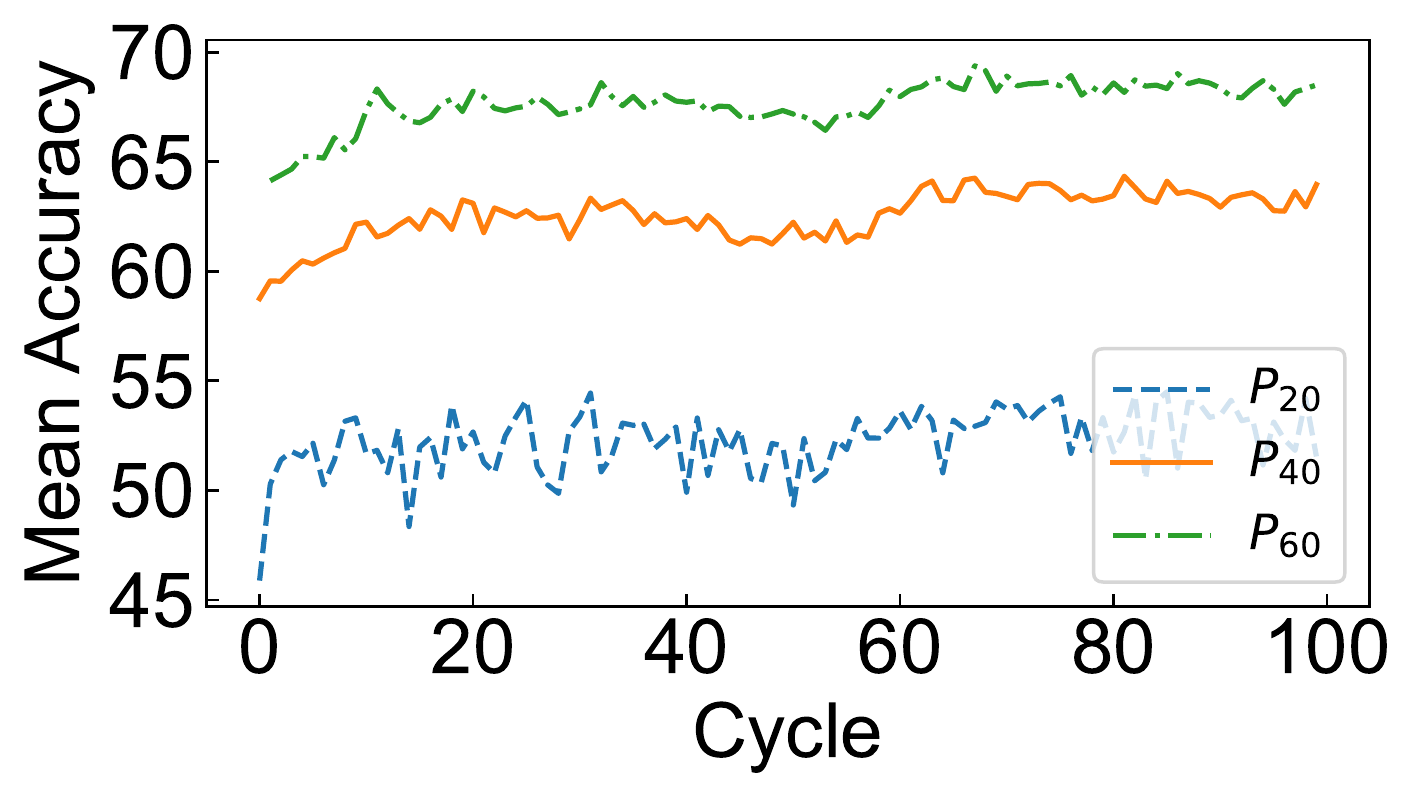}
	\end{center}
	\vspace{-12pt}
	\caption{Average validation accuracy at different cycles for models in $P_{E}, P_{2E}, P_{3E}$, respectively ($E=20$).}
	\label{fig:prog}
	\vspace{6pt}
\end{figure}

\vspace{6pt}\noindent\textbf{Diversity of structures by hierarchical training strategy.}
EcoNAS adopts \emph{hierachical training strategy} to assign newly mutated networks into three $population$ sets and sample networks for mutation from these sets with different probabilities.
Under this scheme, both the good and bad models enjoy chances to mutate so that the architectures in the next evolution cycle will not be trapped in a few local optimal structures. 
We evaluate the average accuracy of networks in the three $poulation$ sets during evolution cycles as shown in Fig.~\ref{fig:prog}.
The difference of average accuracy for networks in $P_{E}$, $P_{2E}$, and $P_{3E}$ are apparent and do not deviate much during evolution cycles, but the average accuracy for networks in those sets increase gradually, 
which verifies the diversity of structures in those $population$ sets. 

The diversity of structures allows the search algorithms to find accurate architectures with fewer search costs by potentially helping to explore the search space more comprehensively.
Since \emph{hierarchical training strategy} provides the diversity of structures in the $population$ sets,
EcoNAS obtains similar competitive architectures after evaluating 1k models.
As a comparison, BlockQNN~\cite{BlockQNN} evaluates 11k models, AmoebaNet~\cite{Amoeba} evaluates 20k models, and NASNet~\cite{TNAS} evaluates 45k models. 

\section{Conclusion}
In this work, we systematically study the behaviors of different reduced settings on maintaining the rank consistency in Neural Architecture Search. 
By conducting extensive experiments on different combinations of reduction factors, we observe that (1) with the same iteration numbers, 
using more training samples with fewer training epochs is more consistent than using more training epochs and fewer training samples;
(2) reducing the resolution of input images is sometimes feasible while reducing the channels of networks is more reliable than reducing the resolution.

We also propose \textit{Economical evolutionary-based NAS (EcoNAS)} that can reduce the search time by about 400$\times$ in comparison to the evolutionary-based state of the art~\cite{Amoeba}.
In EcoNAS, we first design a new, fast, and consistent proxy to accelerate the search process based on the aforementioned observations, 
which also reduces the retraining overhead by retaining more top models.
Then we propose to use a \emph{hierarchical proxy strategy} to assess architectures with different proxy based on their accuracy.
This new strategy improves search efficiency and is capable of finding accurate architectures with less search overhead while exploring the search space more comprehensively.

Last but not least, we find some proxies led by our obeservations are also applicable to other NAS methods~\cite{ProxylessNAS, DARTS}. These proxies further reduce the search time while the discovered models achieve comparable or better performance.
We wish our work could inspire the community to further explore more practical proxies and search algorithms to improve the efficiency of NAS.

\setcounter{section}{0}
\setcounter{figure}{0}
\setcounter{table}{0}
\renewcommand{\thesection}{A\arabic{section}}
\renewcommand{\thetable}{A\arabic{table}}
\renewcommand{\thefigure}{A\arabic{figure}}

\begin{figure*}[t]
   \begin{center}
   \subfigure[Normal Cell]{
   \label{FigA.sub.1}
   \includegraphics[width=0.5\textwidth]{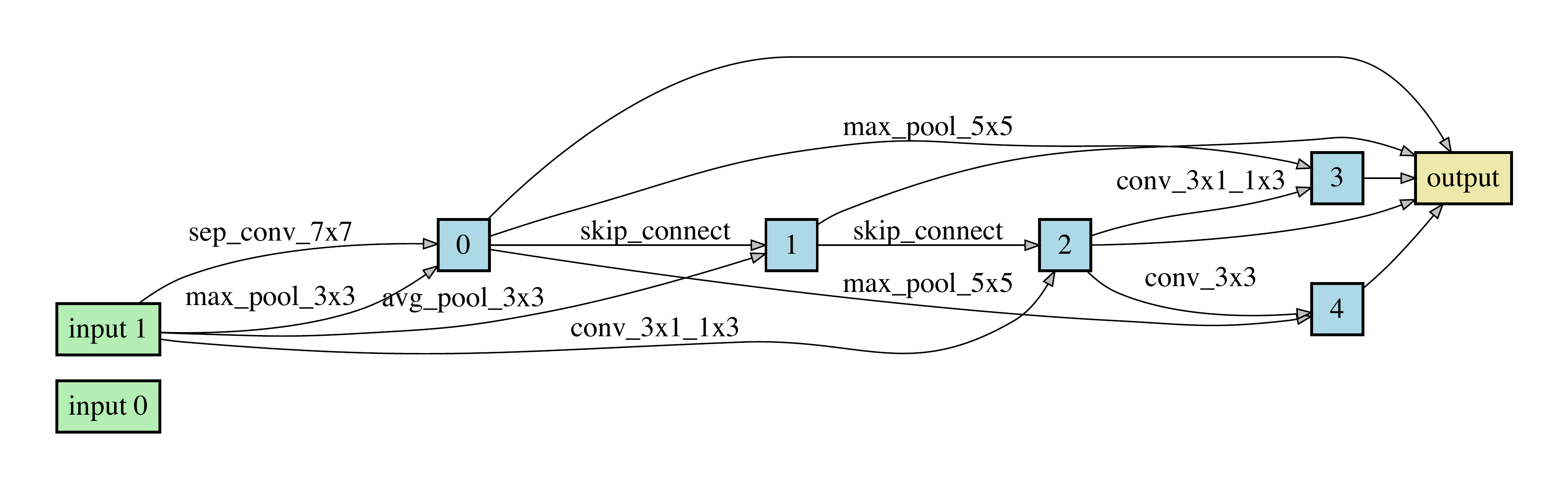}}
   \subfigure[Reduction Cell]{
   \label{FigA.sub.2}
   \includegraphics[width=0.4\textwidth]{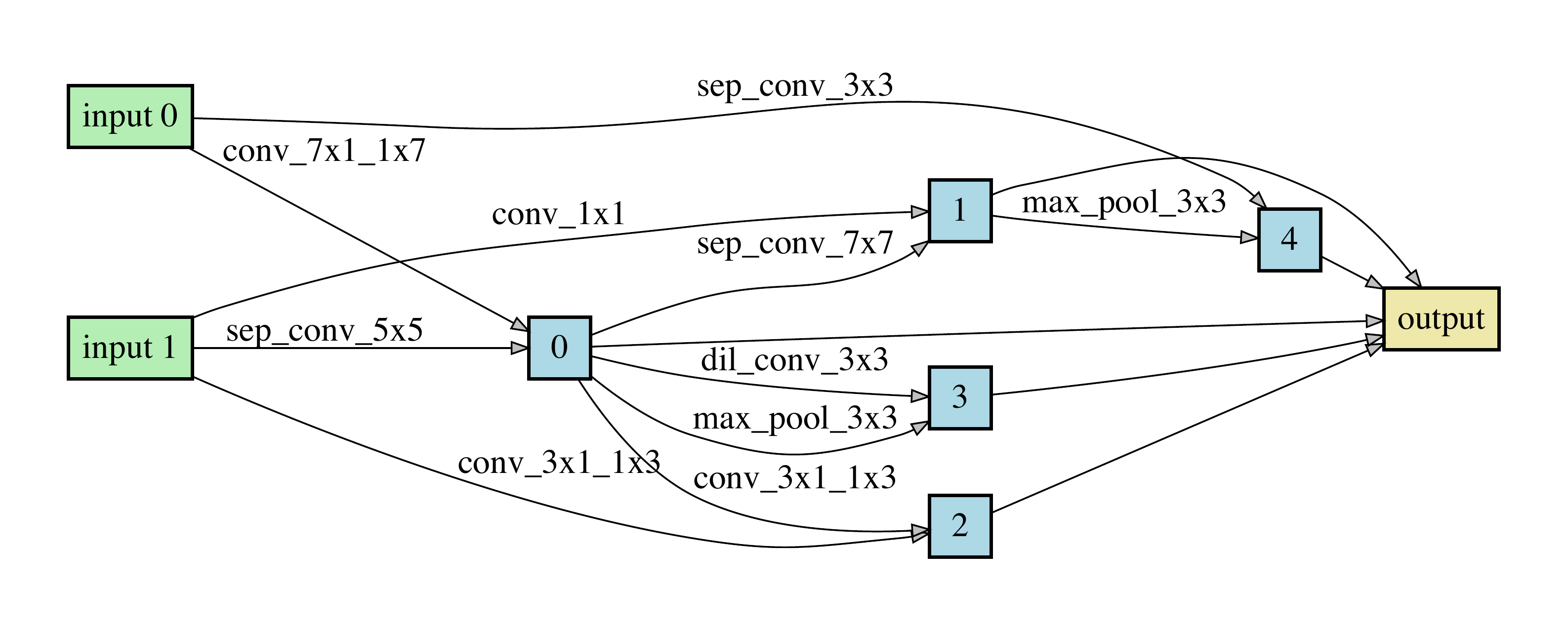}}
   \end{center}
   \caption{Normal and reduction cell structures of model A}
   \label{fig:FigA}
   \end{figure*}
   
   \begin{figure*}[t]
   \begin{center}
   \subfigure[Normal Cell]{
   \label{FigB.sub.1}
   \includegraphics[width=0.45\textwidth]{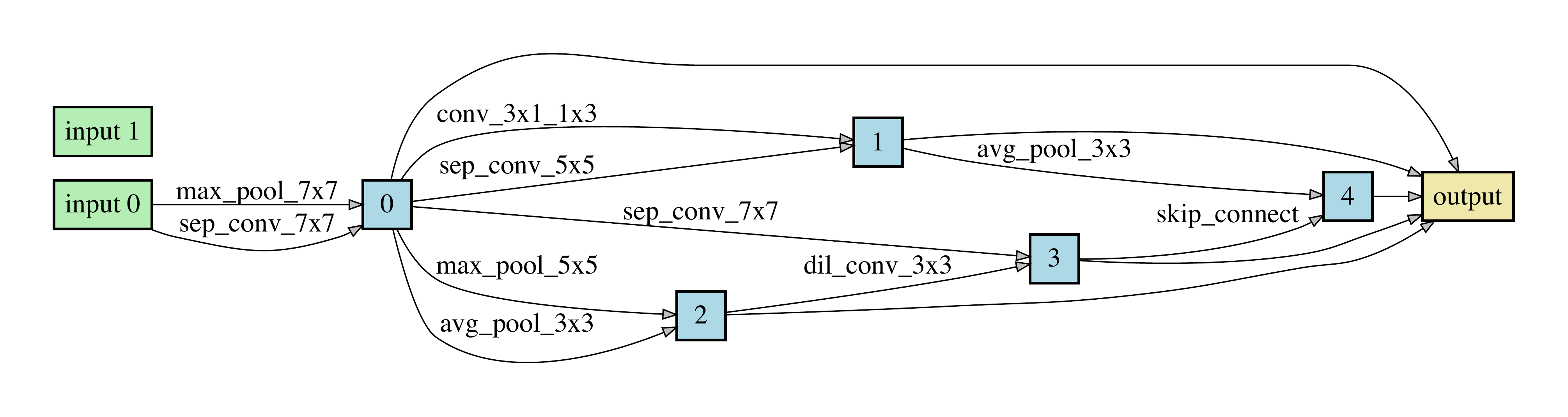}}
   \subfigure[Reduction Cell]{
   \label{FigB.sub.2}
   \includegraphics[width=0.45\textwidth]{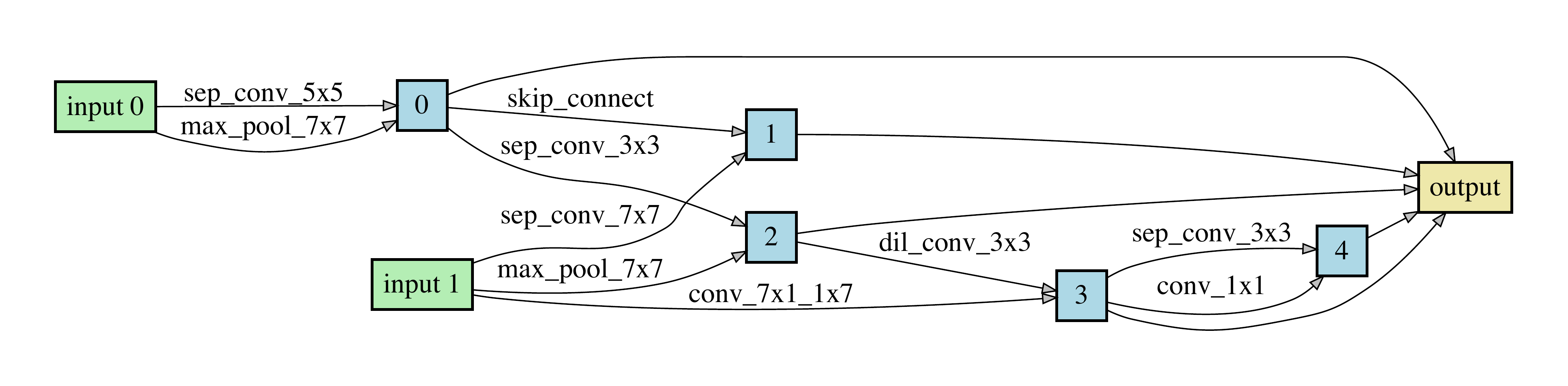}}
   \end{center}
   \caption{Normal and reduction cell structures of model B}
   \label{FigB}
\end{figure*}

\section{Two Model Examples in Introduction}\label{sec:model_example}
In the introduction, we mentioned that some architectures applied with certain reduction factors win in the reduced setting but perform worse in the original setup on CIFAR-10~\cite{CIFAR10}.
The normal and reduction cells for the aforementioned models A and B are shown in Fig.~\ref{fig:FigA} and Fig.~\ref{FigB}, respectively. 
The results on CIFAR-10 of the two models in the original setting and reduced setting are shown in Table \ref{tab:two_models}. 
The original setting is $c_0r_0s_0e_{600}$ while the reduced setting is  $c_0r_0s_0e_{30}$. The training details about the two settings are provided in Section \ref{strat_setting}. 
The results show that the rank of performance evaluated in the reduced setting is not guaranteed to be the same as that in the original setting.

\section{Reliability of Spearman Coefficient}
\begin{figure}[t]
   \begin{center}
   \label{torsp}
   \includegraphics[width=0.45\textwidth]{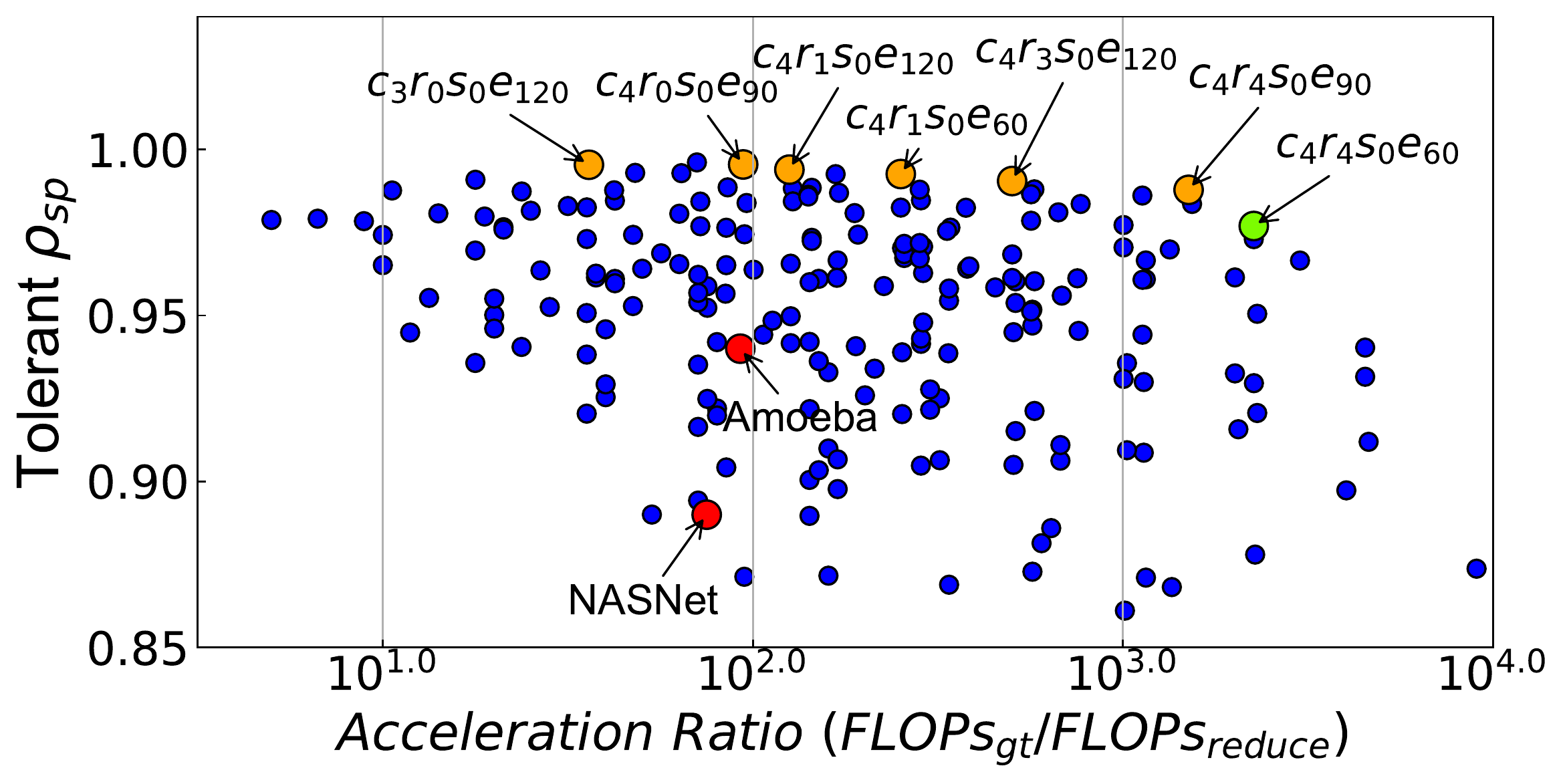}
   \end{center}
   \caption{New $\rho_{sp}$ (Y-axis) and acceleration ratio (X-axis) of reduced settings. Blue points show all settings, the orange ones are good settings and the green one is adopted in EcoNAS.}
\end{figure}

The final accuracy of each network might have minor variation due to the randomness in training. To make the Spearman Coefficient reliable to the accuracy variation, we adjust $\rho_{sp}$ to make it tolerant to the small variations of accuracy and re-analyze the results based on existing records. In the new metric, if the absolute accuracy difference of two models within an interval $b$ are in both original and reduced setting, they will be considered as having no ranking difference. $b$ (0.15\% in our implementation) is used to ignore the minor accuracy variations. For instance, if the accuracy differences of two models are 0.1 in both original and reduced settings, then the new metric will regard these models of having no ranking difference despite the ranking change between these two models. We find that this new $\rho_{sp}$ is highly consistent with previous metric (the normalized correlation is 0.96). And the good settings in Fig.~\ref{fig:recomm} are consistent with the new metric in Fig.~\ref{torsp}. We further test some settings with 100 models and observe consistent results, which also confirms the reliability.

\section{Construction of Model Zoo}\label{model_zoo}
This section provides the details on constructing the model zoo (Section~\ref{sec:exploration}).
Each network architecture in the model zoo is a stack of normal cells alternating with reduction cells.
In each network, these two cells are all generated separately according to the common selection steps in \cite{PNAS, ENAS, TNAS} and we just replace the search algorithm in these approaches by random sampling. 
The number of nodes inside the cells is 5 and every cell receives two initial inputs. 
For cell $k$, the two initial inputs are denoted as $h_{k-2}$ and $h_{k-1}$, which are outputs of previous cells $k-2$ and $k-1$ or the input of images. 
The output of each cell is the depth-wise concatenation of all the intermediate nodes (two initial inputs excluded). 
The generation steps of each intermediate node are as follows:
\begin{itemize}
\item \textbf{Step 1.} Randomly select an input from the input set, which contains two initial inputs of the cell and the set of outputs from previous nodes within the cell.
\item \textbf{Step 2.} Randomly select another input from the same input set as in Step 1.
\item \textbf{Step 3.} Randomly select an operation from the operation set and apply this operation to the first input selected in Step 1.
\item \textbf{Step 4.} Randomly select another operation to apply to the second input selected in Step 2.
\item \textbf{Step 5.} Add the outputs of Step 3 and Step 4 to create the output of the current node.
\end{itemize}

\begin{table}[t]
   \small 
   \caption{The top-1 accuracy on CIFAR-10 for two models in the original setting ($c_0r_0s_0e_{600}$) and the reduced setting ($c_0r_0s_0e_{30}$).}
   \label{tab:two_models}
   \begin{center}
   \begin{tabular}{|c|c|c|}
   \hline
   Model & $c_0r_0s_0e_{600}$ & $c_0r_0s_0e_{30}$ \\
   \hline\hline
   $A$ & 95.27\% & 82.42\% \\
   $B$ & 94.58\% & 86.21\% \\
   \hline
   \end{tabular}
   \end{center}
\end{table}

The original `Step 5' in~\cite{TNAS} provides two combination methods: element-wise addition and depth-wise concatenation. 
However, previous work~\cite{PNAS} mentions that the concatenation method are never chosen during search. Therefore, we only use addition as the combination operation.
We selected 13 operations to build our operation set considering their prevalence in the NAS literature~\cite{ProxylessNAS, DARTS, Amoeba, TNAS}, which are listed as below: 
\vspace{-12pt}\begin{multicols}{2}
\small
\begin{itemize}
    \item 3x3 average pooling
    \item 3x3 max pooling
    \item 5x5 max pooling
    \item 7x7 max pooling
    \item Identity
    \item 1x1 Convolutions
    \item 3x3 Convolutions
    \item 3x3 Separable Convolutions
    \item 5x5 Separable Convolutions
    \item 7x7 Separable Convolutions
    \item 3x3 Dilated Convolutions
    \item 1x3 then 3x1 Convolutions
    \item 1x7 then 7x1 Convolutions
\end{itemize}
\end{multicols}\vspace{-6pt}

\begin{figure*}[t]
   \begin{center}
   \subfigure[Normal Cell]{
   \label{Figbest1.sub.1}
   \includegraphics[width=0.25\textwidth]{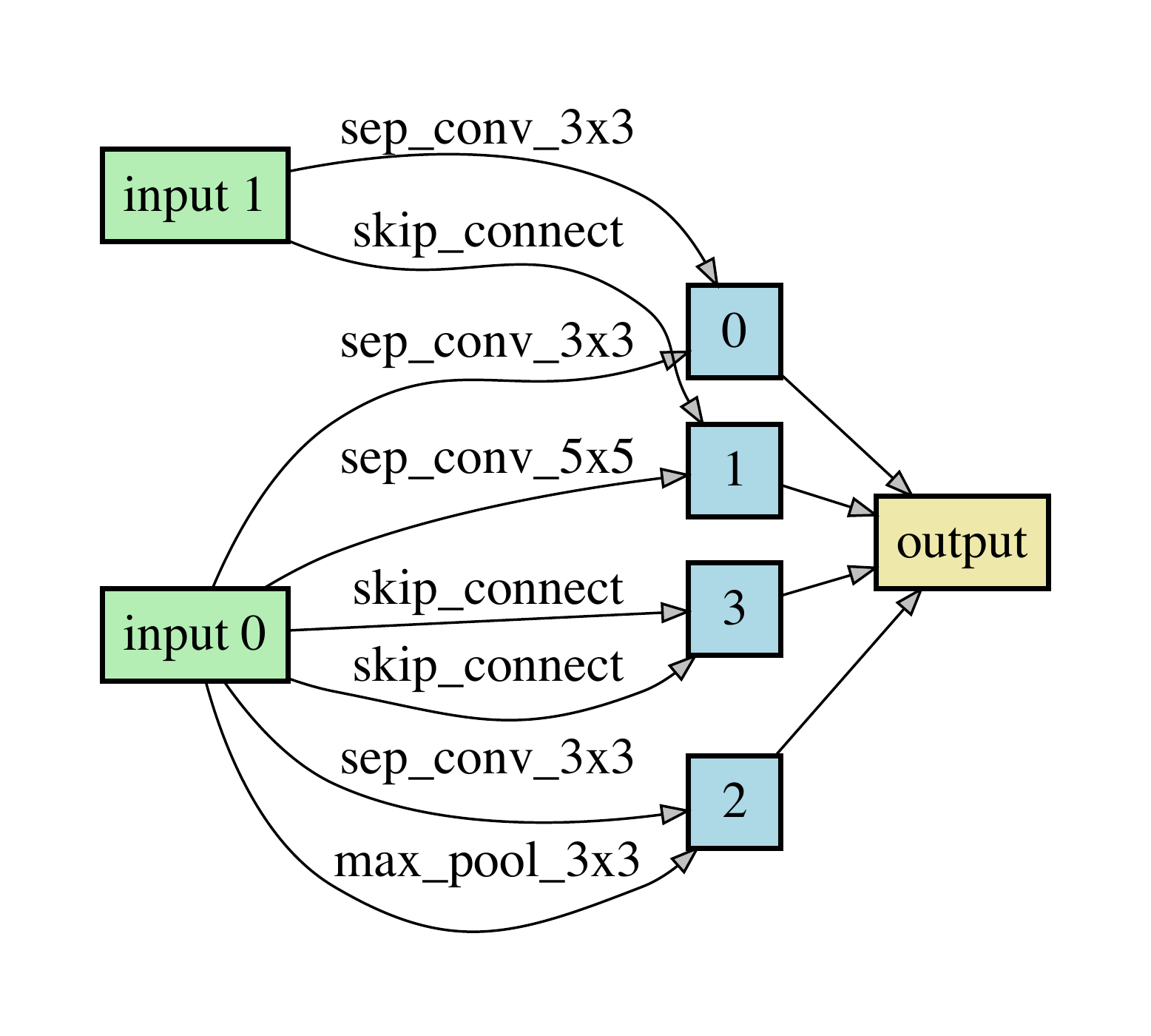}}
   \subfigure[Reduction Cell]{
   \label{Figbest1.sub.2}
   \includegraphics[width=0.6\textwidth]{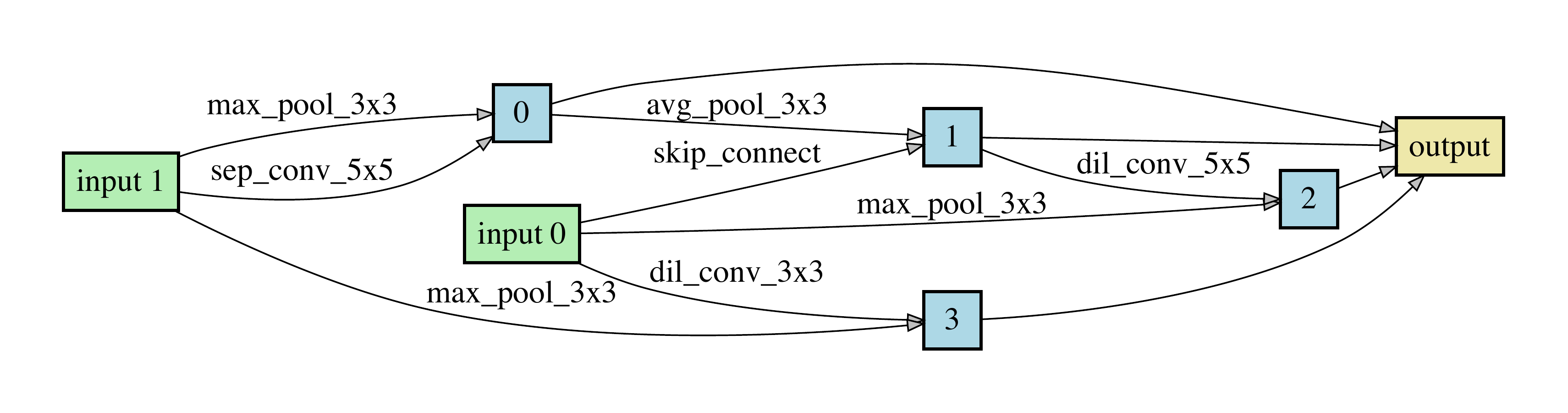}}
   \end{center}
   \vspace{-12pt}
   \caption{Normal and reduction cell structures of first-place model, whose error rate is 2.62\% on CIFAR-10.}
   \label{Figbest1}
\end{figure*}
\begin{figure*}[t]
   \begin{center}
   \subfigure[Normal Cell]{
   \label{Figbest2.sub.1}
   \includegraphics[width=0.25\textwidth]{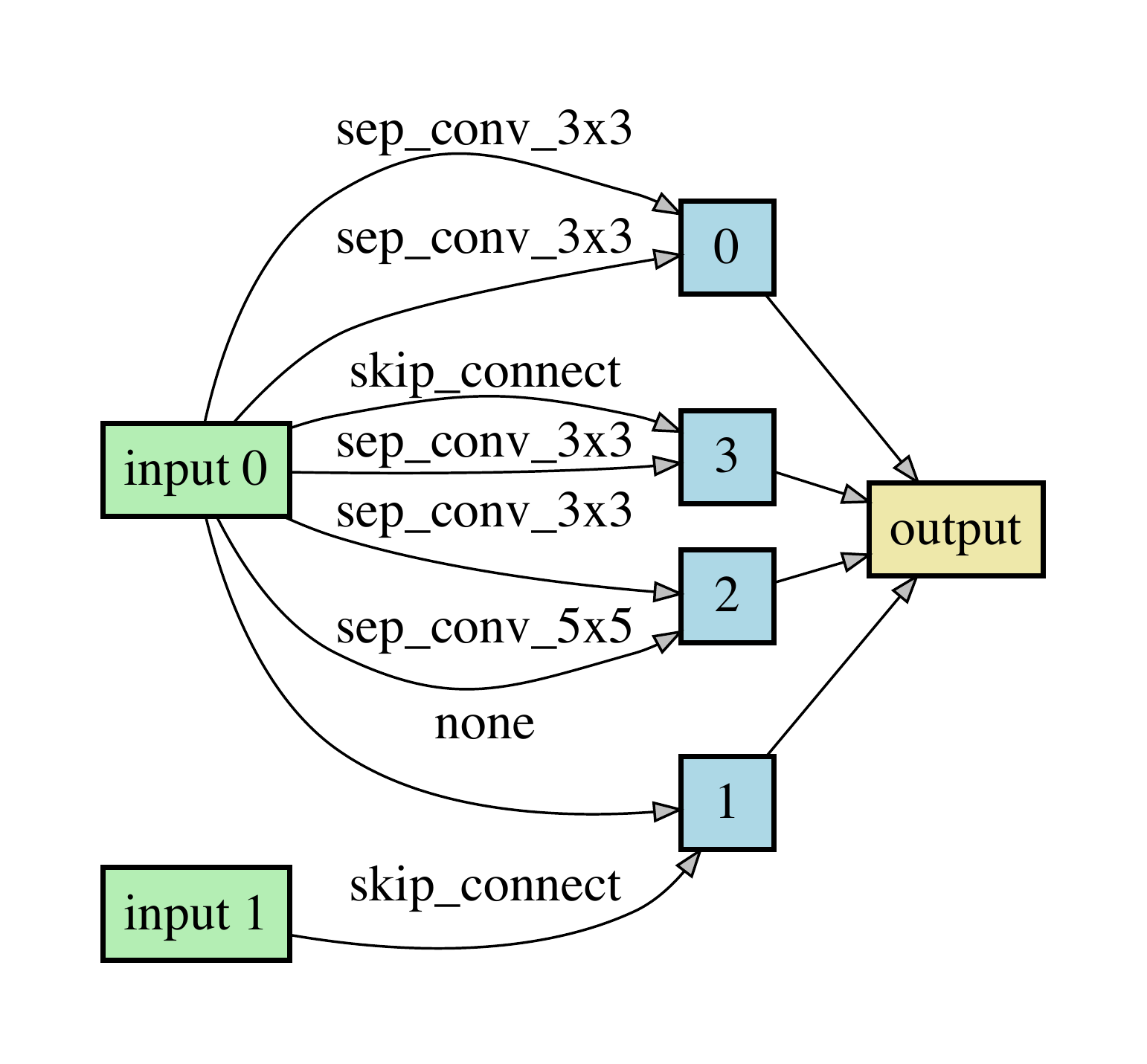}}
   \subfigure[Reduction Cell]{
   \label{Figbest2.sub.2}
   \includegraphics[width=0.45\textwidth]{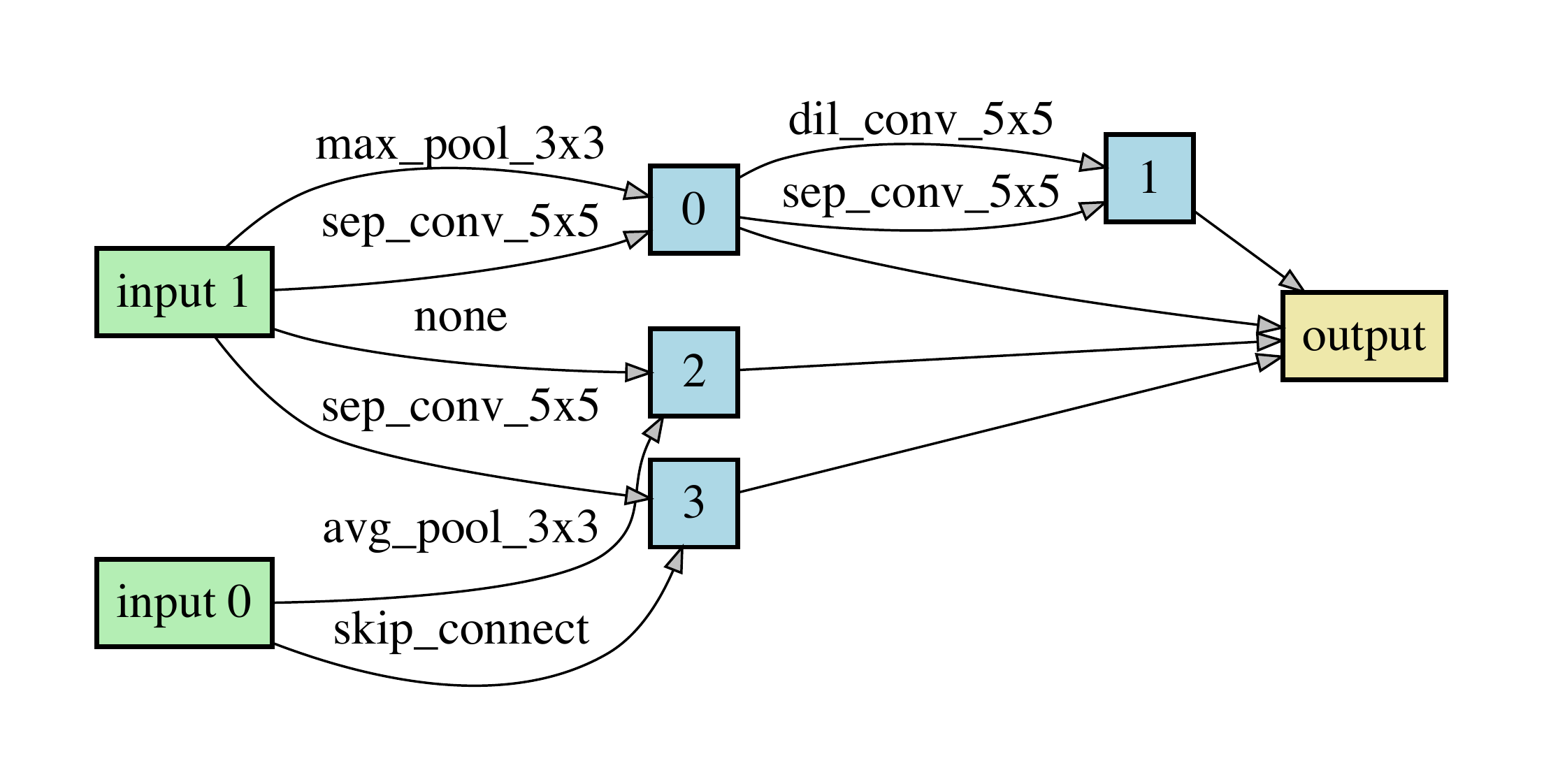}}
   \end{center}
   \vspace{-12pt}
   \caption{Normal and reduction cell structures of second-place model, whose error rate is 2.67\% on CIFAR-10.}
   \label{Figbest2}
\end{figure*}

\section{Detailed Information About \emph{Entropy}}
In Figure~\ref{fig:c_r} and Figure~\ref{fig:ce_re} of Section~\ref{Sec:ExpCIFAR}, a new measurement called \emph{entropy} is used. This section provides the details on how \emph{entropy} is calculated.

We use \emph{entropy}, denoted by $\rho_e$, to measure the monotonicity of a given objective set. 
The \emph{entropy} $\rho_e$ is the Spearman Coefficient measuring the rank difference between the objective set and an arbitrary increasing collection (called base set, such as $\{1,2,3,4,5\}$). 
The objective set is the collection of $\rho_{sp}$ along a certain reduction factor dimension, such as $\rho_{sp}$ of reduced settings $\{c_0r_0s_0e_{30}, c_1r_0s_0e_{30}, c_2r_0s_0e_{30}, c_3r_0s_0e_{30}, c_4r_0s_0e_{30}\}$ along the dimension of reduction factor $c$. 
If the absolute value of $\rho_e$ is closer to 1, it indicates that the objective set has a more apparent increasing ($\rho_e$ approximates $1$) or decreasing ($\rho_e$ approximates $-1$) trend. 
Otherwise (e.g., $\rho_e$ approximates $0$) the monotonicity of the objective set is less apparent.
Since the true values of the inputs will be transferred to the ranks, the choice of base set will not affect the final results if it is a set of increasing numbers.

\section{Experiments}
\subsection{Implementation Details of EcoNAS}
\noindent\textbf{Search space.}
The search space of EcoNAS consists of 8 operations, which follow the previous work~\cite{DARTS} and are listed as follows:
\vspace{-12pt}\begin{multicols}{2}
\small
\begin{itemize}
    \item Zeros
    \item 3x3 average pooling
    \item 3x3 max pooling
    \item 3x3 Separable Convolution
    \item Identity
    \item 5x5 Separable Convolution
    \item 3x3 Dilated Convolution
    \item 5x5 Dilated Convolution
\end{itemize}
\end{multicols}\vspace{-6pt}
Each cell in the network consists of 4 nodes (Figure~\ref{fig:arch}). The generation of each node follows the 5 steps described in Section~\ref{model_zoo}, except that the operation sets are different.
In one cell, the node outputs that are not used will be concatenated together as the cell output~\cite{Amoeba, TNAS}.

\vspace{6pt}\noindent\textbf{Search strategy.}
We use the setting of $c_4 r_4s_0 $ and the batch size is 384. Every network is trained on a single GPU. In every cycle, the chosen networks will be trained for 20 epochs and the maximum training length for each network is 60 epochs, \ie, the complete reduce setting is $c_4 r_4 s_0 e_{60}$, which has been found to be effective in the main text. 
The other hyper-parameters remain the same as stated in Section \ref{strat_setting}. We use $P_{20},P_{40}$ and $P_{60}$ to denote the networks trained for 20, 40, and 60 epochs, respectively. 
In each cycle, 16 networks will be chosen from the population and be mutated, and the top-8 and top-4 networks in $P_{20}$ and $P_{40}$ will continue to be trained for 20 epochs, which means that no more than half of the networks in the $P_{20}$ and $P_{40}$ set will get chance to be continually trained. 
When the process is finished, we only retrain and find the best model from top-5 models from $P_{60}$. 
The searched models that achieve the best and the second best results are shown in Figure~\ref{Figbest1} and Figure~\ref{Figbest2}, respectively.

\subsection{Implementation Details on CIFAR-10}\label{strat_setting}
This section provides the details of training strategies for the original and reduced settings on CIFAR-10 (Section~\ref{Sec:ExpCIFAR} and \ref{sec:cifar_imagenet}).
In the original setting, we train each network from scratch for 600 epochs with batch size of 96. Cosine learning rate schedule is used with $lr_{max}=0.025$ and $lr_{min}=0.001$ and the weight decay is $3e{-4}$~\cite{DARTS}. 
Additional enhancements including cutout~\cite{cutout}, path dropout~\cite{droppath}, and common data augmentations follow the previous work~\cite{DARTS}.

The implementation for the reduced setting follows that for the original setting, except those as follows: 
\begin{packed_enumerate}
\item The number of training epochs is decided by the reduction factor $e$. But the cosine learning rate scheduler still finishes a completed cosine cycle within the reduced epochs.
\item Path dropout is excluded in the reduced setting because we empirically find that the evaluation ability of reduction settings will increase if path dropout is excluded. The possible reason is that we use very small number of epochs, which is not favored by path dropout.
\item The images are resized to reduced resolution after padding and random cropping, and the cutout length is adjusted according to the reduced resolution.
\end{packed_enumerate}

\subsection{Implementation Details on ImageNet}\label{strat_imnet}
This section provides the the details on training strategies for original and reduced settings on ImageNet (Section~\ref{sec:cifar_imagenet}).

In the original setting, the networks are trained for 150 epochs with batch size 2048 on 32 GPUs. The learning rate also follows a cosine annealing schedule with $lr_{max}=0.8$ and $lr_{min}=0.0$. We use warmup \cite{1_hour} to start the learning rate from $0.2$ and then increase it linearly to $0.8$ in the first 2 epochs. The weight decay for all networks is $3e{-5}$. We also use common data augmentation methods following~\cite{He_2016}.

In the reduced setting, the experimental setup is the same as the original setting, except for some differences: (1) The number of training epochs is decided by the reduction factor $e$ and within the reduced epochs, the learning rate schedule will also finish a completed cosine cycle. But we fix the number of warmup epochs to 2 for all the reduced settings. (2) The images will be randomly cropped and then resized to the resolution decided by the reduction factor $r$.


{\small \bibliographystyle{ieee_fullname} \bibliography{sections/egbib}}

\begin{thebibliography}{10}\itemsep=-1pt

\bibitem{15_mins}
Takuya Akiba, Shuji Suzuki, and Keisuke Fukuda.
\newblock Extremely large minibatch {SGD}: {Training} {ResNet-50} on {ImageNet}
  in 15 minutes.
\newblock {\em CoRR}, abs/1711.04325, 2017.

\bibitem{Baker2017AcceleratingNA}
Bowen Baker, Otkrist Gupta, Ramesh Raskar, and Nikhil Naik.
\newblock Accelerating neural architecture search using performance prediction.
\newblock {\em CoRR}, abs/1705.10823, 2017.

\bibitem{Bender2018UnderstandingAS}
Gabriel Bender, Pieter-Jan Kindermans, Barret Zoph, Vijay Vasudevan, and
  Quoc~V. Le.
\newblock Understanding and simplifying one-shot architecture search.
\newblock In {\em ICML}, 2018.

\bibitem{Brock2017SMASHOM}
Andrew Brock, Theodore Lim, James~M. Ritchie, and Nick Weston.
\newblock {SMASH}: One-shot model architecture search through hypernetworks.
\newblock {\em CoRR}, abs/1708.05344, 2017.

\bibitem{ProxylessNAS}
Han Cai, Ligeng Zhu, and Song Han.
\newblock Proxylessnas: Direct neural architecture search on target task and
  hardware.
\newblock In {\em ICLR}, 2019.

\bibitem{Chen2019DetNASBS}
Yukang Chen, Tong~Rui Yang, Xiangyu Zhang, Gaofeng Meng, Chunhong Pan, and Jian
  Sun.
\newblock {DetNAS}: {Backbone} search for object detection.
\newblock In {\em NeurIPS}, 2019.

\bibitem{peephole}
Boyang Deng, Junjie Yan, and Dahua Lin.
\newblock Peephole: Predicting network performance before training.
\newblock {\em CoRR}, abs/1712.03351, 2017.

\bibitem{cutout}
Terrance DeVries and Graham~W. Taylor.
\newblock Improved regularization of convolutional neural networks with cutout.
\newblock {\em CoRR}, abs/1708.04552, 2017.

\bibitem{nas_fpn}
Golnaz Ghiasi, Tsung{-}Yi Lin, Ruoming Pang, and Quoc~V. Le.
\newblock {NAS-FPN:} learning scalable feature pyramid architecture for object
  detection.
\newblock {\em CoRR}, abs/1904.07392, 2019.

\bibitem{1_hour}
Priya Goyal, Piotr Dollár, Ross Girshick, Pieter Noordhuis, Lukasz Wesolowski,
  Aapo Kyrola, Andrew Tulloch, Yangqing Jia, and Kaiming He.
\newblock Accurate, large minibatch sgd: Training imagenet in 1 hour.
\newblock {\em CoRR}, abs/1706.02677, 2017.

\bibitem{guo2019single}
Zichao Guo, Xiangyu Zhang, Haoyuan Mu, Wen Heng, Zechun Liu, Yichen Wei, and
  Jian Sun.
\newblock Single path one-shot neural architecture search with uniform
  sampling.
\newblock {\em CoRR}, abs/1904.00420, 2019.

\bibitem{He_2016}
Kaiming He, Xiangyu Zhang, Shaoqing Ren, and Jian Sun.
\newblock Deep residual learning for image recognition.
\newblock In {\em CVPR}, Jun 2016.

\bibitem{mobilenet_v3}
Andrew Howard, Mark Sandler, Grace Chu, Liang{-}Chieh Chen, Bo Chen, Mingxing
  Tan, Weijun Wang, Yukun Zhu, Ruoming Pang, Vijay Vasudevan, Quoc~V. Le, and
  Hartwig Adam.
\newblock Searching for {MobileNetV3}.
\newblock {\em CoRR}, abs/1905.02244, 2019.

\bibitem{DenseNet}
Gao Huang, Zhuang Liu, Laurens van~der Maaten, and Kilian~Q. Weinberger.
\newblock Densely connected convolutional networks.
\newblock In {\em CVPR}, 2017.

\bibitem{CIFAR10}
Alex Krizhevsky.
\newblock Learning multiple layers of features from tiny images, 2009.
\newblock Technical report,.

\bibitem{droppath}
Gustav Larsson, Michael Maire, and Gregory Shakhnarovich.
\newblock Fractalnet: Ultra-deep neural networks without residuals.
\newblock In {\em ICLR}, 2017.

\bibitem{Liu_2019_CVPR}
Chenxi Liu, Liang-Chieh Chen, Florian Schroff, Hartwig Adam, Wei Hua, Alan~L.
  Yuille, and Li Fei-Fei.
\newblock {Auto-DeepLab}: {Hierarchical} neural architecture search for
  semantic image segmentation.
\newblock In {\em CVPR}, 2019.

\bibitem{PNAS}
Chenxi Liu, Barret Zoph, Maxim Neumann, Jonathon Shlens, Wei Hua, Li-Jia Li, Li
  Fei-Fei, Alan Yuille, Jonathan Huang, and Kevin Murphy.
\newblock Progressive neural architecture search.
\newblock In {\em ECCV}, 2018.

\bibitem{hier_evo}
Hanxiao Liu, Karen Simonyan, Oriol Vinyals, Chrisantha Fernando, and Koray
  Kavukcuoglu.
\newblock Hierarchical representations for efficient architecture search.
\newblock {\em CoRR}, abs/1711.00436, 2017.

\bibitem{DARTS}
Hanxiao Liu, Karen Simonyan, and Yiming Yang.
\newblock {DARTS}: Differentiable architecture search.
\newblock In {\em ICLR}, 2019.

\bibitem{CNN_adv}
Dmytro Mishkin, Nikolay Sergievskiy, and Jiri Matasa.
\newblock Systematic evaluation of cnn advances on the imagenet.
\newblock {\em CVIU}, 2017.

\bibitem{ENAS}
Hieu Pham, Melody~Y. Guan, Barret Zoph, Quoc~V. Le, and Jeff Dean.
\newblock Efficient neural architecture search via parameter sharing.
\newblock In {\em ICML}, 2018.

\bibitem{Amoeba}
Esteban Real, Alok Aggarwal, Yanping Huang, and Quoc~V. Le.
\newblock Regularized evolution for image classifier architecture search.
\newblock In {\em AAAI}, 2019.

\bibitem{Real2017LargeScaleEO}
Esteban Real, Sherry Moore, Andrew Selle, Saurabh Saxena, Yutaka~Leon Suematsu,
  Jie Tan, Quoc~V. Le, and Alexey Kurakin.
\newblock Large-scale evolution of image classifiers.
\newblock In {\em ICML}, 2017.

\bibitem{ILSVRC15}
Olga Russakovsky, Jia Deng, Hao Su, Jonathan Krause, Sanjeev Satheesh, Sean Ma,
  Zhiheng Huang, Andrej Karpathy, Aditya Khosla, Michael Bernstein,
  Alexander~C. Berg, and Li Fei-Fei.
\newblock {ImageNet Large Scale Visual Recognition Challenge}.
\newblock {\em IJCV}, 2015.

\bibitem{tan2018mnasnet}
Mingxing Tan, Bo Chen, Ruoming Pang, Vijay Vasudevan, and Quoc~V. Le.
\newblock {MnasNet: Platform}-aware neural architecture search for mobile.
\newblock {\em CoRR}, abs/1807.11626, 2018.

\bibitem{All_Images}
Kailas Vodrahalli, Ke Li, and Jitendra Malik.
\newblock Are all training examples created equal? an empirical study.
\newblock {\em CoRR}, abs/1811.12569, 2018.

\bibitem{wang2019nasfcos}
Ning Wang, Yang Gao, Hao Chen, Peng Wang, Zhi Tian, and Chunhua Shen.
\newblock {NAS-FCOS}: Fast neural architecture search for object detection.
\newblock {\em CoRR}, abs/1906.04423, 2019.

\bibitem{FBNet}
Bichen Wu, Xiaoliang Dai, Peizhao Zhang, Yanghan Wang, Fei Sun, Yiming Wu,
  Yuandong Tian, Peter Vajda, Yangqing Jia, and Kurt Keutzer.
\newblock {FBNet}: Hardware-aware efficient convnet design via differentiable
  neural architecture search.
\newblock In {\em CVPR}, 2019.

\bibitem{SNAS}
Sirui Xie, Hehui Zheng, Chunxiao Liu, and Liang Lin.
\newblock {SNAS}: Stochastic neural architecture search.
\newblock In {\em ICLR}, 2019.

\bibitem{large_bs_training}
Yang You, Igor Gitman, and Boris Ginsburg.
\newblock Large batch training of convolutional networks.
\newblock {\em CoRR}, abs/1708.03888v3, 2017.

\bibitem{mixup}
Hongyi Zhang, Moustapha Cisse, Yann~N. Dauphin, and David Lopez-Paz.
\newblock mixup: Beyond empirical risk minimization.
\newblock In {\em ICLR}, 2018.

\bibitem{BlockQNN}
Zhao Zhong, Junjie Yan, Wei Wu, Jing Shao, and Cheng-Lin Liu.
\newblock Practical block-wise neural network architecture generation.
\newblock In {\em CVPR}, 2018.

\bibitem{NAS}
Barret Zoph and Quoc~V. Le.
\newblock Neural architecture search with reinforcement learning.
\newblock In {\em ICLR}, 2017.

\bibitem{TNAS}
Barret Zoph, Vijay Vasudevan, Jonathon Shlens, and Quoc~V. Le.
\newblock Learning transferable architectures for scalable image recognition.
\newblock In {\em CVPR}, 2018.

\end{thebibliography}

\end{document}